\documentclass{article}

\usepackage{microtype}
\usepackage{graphicx}
\usepackage{booktabs} 

\usepackage{hyperref}

\usepackage[accepted]{sysml2019}

\usepackage{xspace}
\usepackage{xcolor}
\usepackage{graphicx}
\usepackage{booktabs}
\usepackage{varwidth}
\usepackage{capt-of}
\usepackage{amssymb}
\usepackage{subcaption}
\newcommand{\sysname}{SimEx\xspace} 
\DeclareMathAlphabet{\mathbfcal}{OMS}{cmsy}{b}{n}

\sysmltitlerunning{SimEx: Express Prediction of Inter-dataset Similarity by a Fleet of Autoencoders}

\begin{document}

\twocolumn[
\sysmltitle{SimEx: Express Prediction of Inter-dataset Similarity by a Fleet of Autoencoders}



\sysmlsetsymbol{equal}{*}

\begin{sysmlauthorlist}
\sysmlauthor{Inseok Hwang}{ibm}
\sysmlauthor{Jinho Lee}{yonsei}
\sysmlauthor{Frank Liu}{oakridge}
\sysmlauthor{Minsik Cho}{ibm}
\end{sysmlauthorlist}

\sysmlaffiliation{ibm}{IBM, Austin, Texas, USA}
\sysmlaffiliation{oakridge}{Oak Ridge National Laboratory, Oak Ridge, Tennessee, USA}
\sysmlaffiliation{yonsei}{Yonsei University, Seoul, Korea}

\sysmlcorrespondingauthor{Inseok Hwang}{ihwang@us.ibm.com}

\sysmlkeywords{Machine Learning, SysML}

\vskip 0.3in

\begin{abstract}
Knowing the similarity between sets of data has a number of positive implications in training an effective model, such as assisting an informed selection out of known datasets favorable to model transfer or data augmentation problems with an unknown dataset. 
Common practices to estimate the similarity between data include comparing in the original sample space, comparing in the embedding space from a model performing a certain task, or fine-tuning a pretrained model with different datasets and evaluating the performance changes therefrom. 
However, these practices would suffer from shallow comparisons, task-specific biases, or extensive time and computations required to perform comparisons. 
We present \sysname, a new method for early prediction of inter-dataset similarity using a set of pretrained autoencoders each of which is dedicated to reconstructing a specific part of known data. 
Specifically, our method takes unknown data samples as input to those pretrained autoencoders, and evaluate the difference between the reconstructed output samples against their original input samples. 
Our intuition is that, the more similarity exists between the unknown data samples and the part of known data that an autoencoder was trained with, the better chances there could be that this autoencoder makes use of its trained knowledge, reconstructing output samples closer to the originals. 
We demonstrate that our method achieves more than 10x speed-up in predicting inter-dataset similarity compared to common similarity-estimating practices. We also demonstrate that the inter-dataset similarity estimated by our method is well-correlated with common practices and outperforms the baselines approaches of comparing at sample- or embedding-spaces, without newly training anything at the comparison time. 

\end{abstract}
]



\printAffiliationsAndNotice{}  

\section{Introduction}

We have been witnessing a continuing proliferation of new datasets, largely spurred by the record-breaking success of deep neural networks and the ubiquity of data-generation and sharing tools. 
In spite of such an abundance of datasets, having the right dataset well-suited for a target application is not guaranteed in practice. 
The performance of a machine learning model is largely dependent on the availability of a relevant, adequate, and balanced dataset that well represents the distribution of the application-specific sample space. 
However, it is quite often that real-world applications accompany small-sized or poorly organized data of their own. 
Certain practices are commonly exercised, such as transferring from a model pretrained with another dataset or augmenting the given training data with samples from other datasets~\cite{bengio2011deep, DBLP:journals/corr/abs-1003-0358, dai2007boosting, ge2017borrowing, perez2017effectiveness, seltzer2013investigation, yosinski2014transferable}.
Still, it is often not obvious to foresee the compatibility of one of the available known datasets with respect to the target sample space. 

In this paper, we present \textit{\sysname}, a new method for early prediction of inter-dataset similarity. 
Specifically, \sysname takes unknown data samples as input to a set of autoencoders each of which was pretrained to reconstruct a specific, distinct part of known data. Then, the differences between the input samples and the reconstructed output samples are evaluated. 
Our intuition is that, the more underlying similarity the unknown data samples share with the specific part of known data that an autoencoder was trained with, the better chances there could be that this autoencoder makes use of its learned knowledge, reconstructing output samples closer to the originals. Here, the differences between the original inputs and the reconstructed outputs constitute a relative indicator of similarity between the unknown data samples and the specific part of known data.

\sysname implies a number of practical benefits. 
Empirical evidences support that the similarity between data is correlated with the effectiveness of transferring a pretrained network~\cite{yosinski2014transferable} or supplementing the training samples from a relevant dataset~\cite{dai2007boosting, ge2017borrowing}. 
Not only this fundamental benefit, the properties in how \sysname predicts the data similarity could lead to further advantages. 
It is likely that \sysname takes account of self-organized deep features about the given data, beyond the sample space measures such as pixel-level distribution~\cite{fuglede2004jensen, kullback1951information} or structure-level similarity~\cite{oliva2001modeling, wang2004image}.
Notably, despite potential use of deep features, \sysname does not measure a difference metric directly from the latent space; it takes the measurement back to the sample space where the reconstructed output appears. 
An implication from this property is that it may help alleviate possible model biases to the particular way that a model represents the deep features. 
Another benefit is from the systems perspectives. 
At comparison time, \sysname predicting the similarity with respect to an unknown dataset does not require any further training, because the autoencoders were pretrained.
This property saves a considerable amount of runtime resources at comparison time, unlike the existing practice of inferring the relevance between datasets by transferring a network from one dataset to another and measuring the resulting performance deviation~\cite{yosinski2014transferable}.
Our experiments show that \sysname is more than 10 times faster than the existing transfer learning-based methods at runtime to predict the similarity between an unknown dataset and reference datasets.

We note that the term `similarity' is not a single rigorously defined measure. Various similarity measures have been defined, each with a target-specific focus~\cite{larsen2016autoencoding}. 
Yet, we believe that a similarity measure making little assumption on the data or task would be favorable to wide usability. 
In this paper, we do not claim that \sysname reflects a `true' similarity of any kind. Instead, this paper focuses on experimental exploration of the usability and benefits of \sysname-predicted similarity in the context of typical transfer learning and data augmentation. 

Our contributions are threefold. First, we present a new method predicting the similarity between data, which is essentially: have a set of autoencoders learn about known data, represent unknown data with respect to those learned models, reconstruct the unknown data from that representation, and measure the reconstruction errors in the sample space which is considered an indicator of the relative similarity of the unknown data with respect to the known data. 
Second, we devise applying our method to three cases of similarity prediction: the inter-dataset similarity, the inter-class similarity across heterogeneous datasets, and the inter-class similarity within a single dataset. 
Third, we demonstrate the clear speed advantage and potential usability of our method in making informed decisions in the practical problems: transferring pretrained networks, augmenting a small dataset for a classification task, and estimating the inter-class confusion levels in a labeled dataset. 

\section{Related Works}

Quantifying the similarity between two different datasets is a well-studied topic in machine learning. A theoretical abstraction of data similarity can be cast to the classic KL-divergence~\cite{kullback1951information}. For `shallow' datasets, empirical metrics such as Maximal Mean Discrepancy (MMD)~\cite{borgwardt2006integrating} are popular choices. 
For images, the structural similarity metric (SSIM) is a well-known metric taking account of luminance, contrast, and structural information~\cite{wang2004image}.
However, for learning with high-dimensional data such as complex visual applications, it is challenging to directly apply these shallow methods.

A related topic is {\em domain adaptation}; a model trained on the dataset in the {\em source} domain is extended to carry out the task on the data in the {\em target} domain. For shallow features, techniques such as Geodesic Flow Sampling (GFS)~\cite{gopalan2013unsupervised}, Geodesic Flow Kernel (GFK)~\cite{gong2012geodesic}, and Reproducing Kernel Hilbert Space (RKHS)\cite{ni2013subspace} are well-established. Deep methods have been proposed recently, such as Domain Adversarial Neural Networks (DANN)~\cite{ganin2016domain}, Adversarial Discriminative Domain Adaptation (ADDA)~\cite{tzeng2017adversarial} and Deep Adaptation Network (DAN)~\cite{long2015learning}. However, the goal of domain adaptation is fundamentally different from data similarity.  Our objective is to directly compute the data similarity without being tied to a specific machine learning method or a specific neural network architecture. 

A problem space relevant to data similarity is to improve the output image quality of generative models, such as by having the models incorporate the structural knowledge~\cite{snell2017learning, yan2016attribute2image} or perform intelligent sharpening~\cite{mansimov2015generating}. 
For \sysname, however, more accurate reconstruction of a source sample is preferable but not imperative. 
The primary interest of \sysname is to discern the \textit{relative} differences of reconstruction quality between input samples.
For this purposes, modest reconstruction quality is still acceptable and may be even favored considering the extensive system resources likely to be consumed to achieve high-quality reconstruction.

\section{Scenarios} \label{sec:scenarios}
In this section, we list a few practical scenarios where our similarity prediction using \sysname could be potentially beneficial in solving some real-world problems. 
In each scenario, \sysname is used in slightly different ways, but under the same methodology.

\subsection{Inter-dataset similarity}
Transferring from an existing classifier pretrained with a relevant dataset is frequently exercised to obtain a good classifier for a new dataset, especially when the new dataset is small. 
Conversely, the accuracy deviation of the transferred model is often an implicit indicator of the similarity of the new dataset with respect to the known dataset~\cite{yosinski2014transferable}. 

Suppose a service provider who trains classifiers for small datasets coming from a number of customers.
The service provider would possess in its library a set of classifiers pretrained from various datasets.
Upon receiving the customer dataset, transfer learning from one of the pretrained classifier can be performed, to learn a new classifier in a short amount of time, and from small number of samples.

However, choosing the right model to transfer from is a non-trivial problem. 
Intuitively, it makes sense to choose the dataset that is the most `similar' to the target dataset.
The best practice is to try all the classifiers in the library and choose the best, at the expense of the high computing/time cost each time a transfer learning is performed.

\sysname-predicted similarity between the datasets can be used as a proxy to the which classifier will be the best for transfer learning.
Later in this paper, we will show that \sysname can achieve a meaningfully consistent result compared to the actual transferred quality.
At the same time, \sysname achieves more than an order of magnitude faster runtime latency since it does not involve training.

\subsection{Inter-class similarity across heterogeneous datasets}
It is a common data augmentation strategy to supplement a given set of data with relevant existing, deformed, or synthesized samples~\cite{DBLP:journals/corr/abs-1003-0358, perez2017effectiveness, seltzer2013investigation}. 
If the task is classification, it is logical to supplement each class with samples that are highly relevant to that class. 
Suppose that we have many `reference' datasets already labeled, potentially some of which might be relevant to a new target classification problem concerning a new, insufficient dataset. 

Hence it is crucial to answer the following questions: which existing dataset is most beneficial to strengthen the target classification model? and further, which classes of the existing datasets are the most beneficial to each target class?

We believe this falls into where \sysname can provide some information, 
by predicting similarities between classes among different datasets.
For example, if we have a `food' class in the target dataset, we can try to see which one among `flower' class or `fish' class is better for supplementing.
Again, \sysname saves the huge runtime cost of trial-and-error method, where each target class is supplemented with arbitrary or hand-picked reference class and the accuracy is checked after the training.

\subsection{Inter-class similarity within a single dataset}
Classification is a very common kind of machine learning problems. 
Many conventional applications utilizing classifiers would be interested in only a single class label of the highest softmax output. 
But it is known that the classifiers are not equally confident of every class, but exhibit varying levels of confusion between different input-output class pairs~\cite{delahunt2019money}. 

Knowing potential inter-class confusion in advance would bring practical benefits in real-world problems. 
Real datasets may have ill-labeled samples in some classes. Even some classes may be suboptimally separated at the first place. 
For example, a dataset of fruit classes \{apple, banana, kiwi, clementine, tangerine\} would yield higher confusion between two citrus classes. 
One may also want to know potential inter-class confusion when a new class has been introduced to an existing classifier. Thus, when training a classifier with a dataset of unknown characteristics, early screening of potential inter-class confusion may help make informed actions, e.g., re-clustering problematic classes, before training the classifier with the given dataset as-is. 
It could save considerable time and resources invested to possible trials and errors. 

We can hypothesize that classifiers would suffer from more confusion among `similar' classes, and \sysname can give a rough forecast on it by predicting the similarities between the classes.

\begin{figure*}[ht]

    \begin{subfigure}{0.25\linewidth}
        \centering 
        \includegraphics[ width=0.90\linewidth]{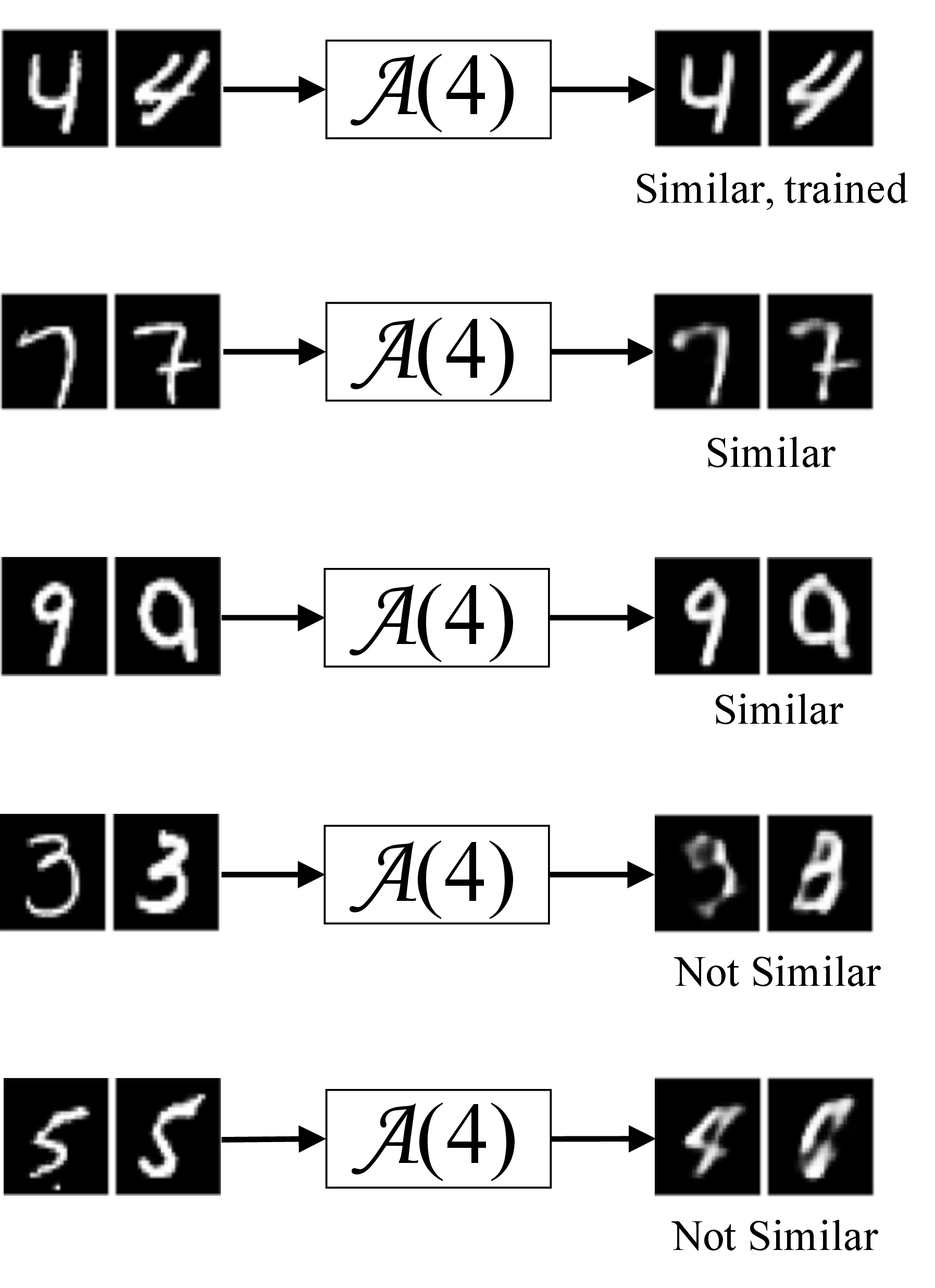}
        \caption{Reconstructing various input digits by the autoencoder $\mathcal{A}(4)$ trained with samples of digit 4. }
        \label{fig:reconstructed_digits}
    \end{subfigure}
    \begin{subfigure}{0.74\linewidth}
        \centering
        \includegraphics[trim=0 0 0mm 0, clip=true, width=0.99\linewidth]{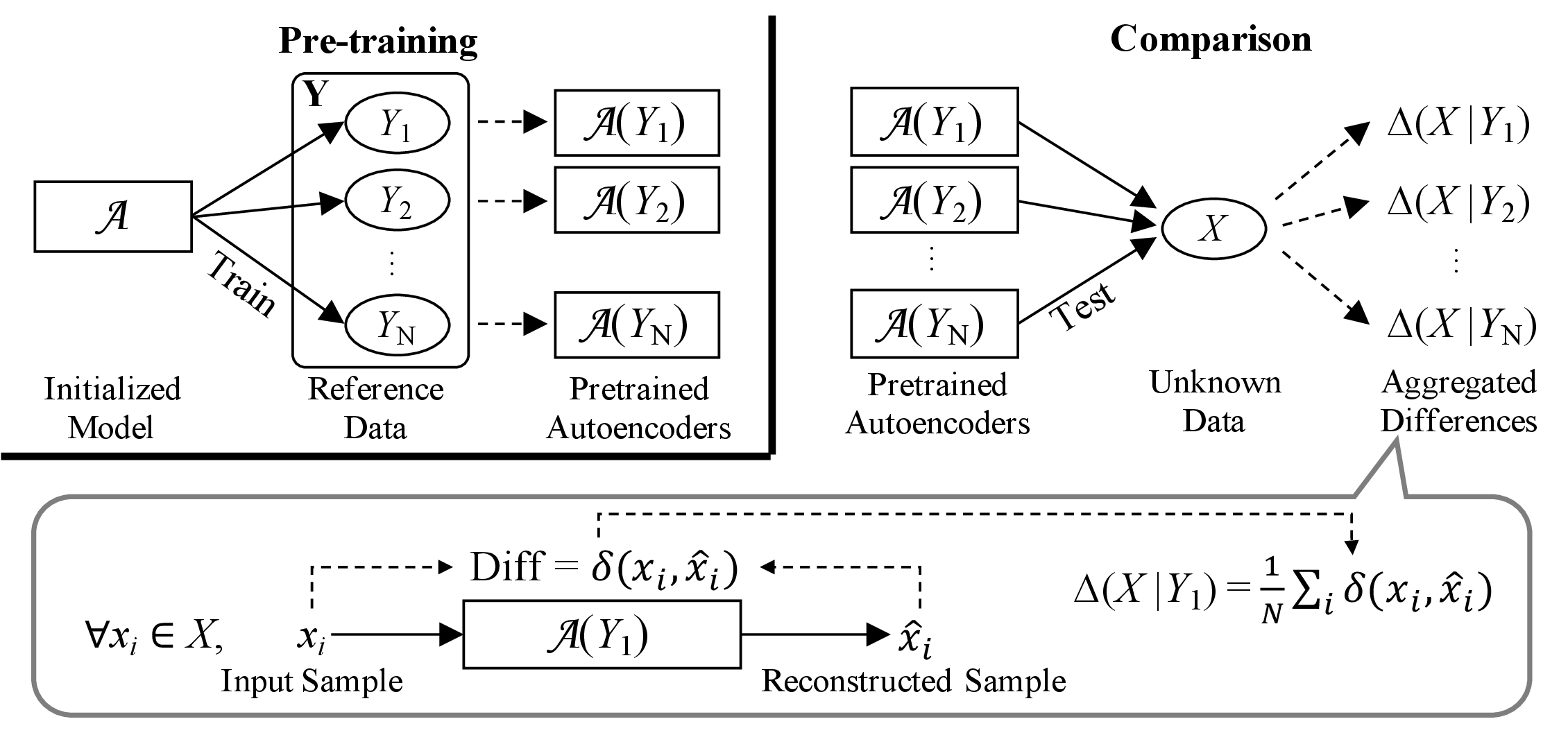}
        \caption{Diagrams of pretraining (upper-left) and performing comparisons.}
        \label{fig:operation_diagram}
    \end{subfigure}
    \caption{Layer specifications, operation diagrams, and reconstruction examples. }
\end{figure*}

\section{\sysname Overview} \label{sec:methods}

Throughout this paper, we use the notation $\mathcal{A}(X)$ to denote an autoencoder pretrained with samples $x_i \in X$.
Suppose that we have a set of known data samples readily available $\mathbf{Y}$, which is a union of smaller disjoint parts of data: $\mathbf{Y} = Y_0 \cup Y_1 \cup ... \cup Y_N$. 
In practice, $\mathbf{Y}$ could be a union of independent datasets $Y_i$, or without loss of generality, $\mathbf{Y}$ could be a labeled dataset consisting of classes $Y_i$. 
Our methods utilize a set of autoencoders $\mathbfcal{A}=\{\mathcal{A}(Y_1), \mathcal{A}(Y_2), ..., \mathcal{A}(Y_N)\}$. 
Each $\mathcal{A}(Y_i)$ is specialized in reconstructing a specific part of the data space (i.e., $Y_i$ out of the whole $\mathbf{Y}$). 

In this section, we present the key properties of our autoencoders, and their implications in predicting the similarity-order of multiple sets of data with respect to a reference set of data. 
Figure~\ref{fig:reconstructed_digits} illustrates examples from an autoencoder trained to reconstruct the samples of digit 4. 
For each row, the left shows the input samples and the right shows the corresponding output samples. 
The topmost row demonstrates the results from the input samples of digit 4 taken from the test set. 
Not surprisingly, each output sample resembles its input sample very closely. 
The 2nd and 3rd rows show the examples with digit 7 and 9, respectively; their reconstructions look slightly degraded compared to the examples with digit 4, but the reconstructions look still fairly close.  
In contrast, the 4th and 5th rows, showing the examples with digit 3 and 5, exhibit severely degraded results in both samples.
Here we can hypothesize that the knowledge learned from reconstructing 4 has been useful for reconstructing 7 and 9 than for 3 and 5 and that 4 could be thought as to be closer to 7 and 9.

Figure~\ref{fig:operation_diagram} illustrates the diagrams of pretraining and performing comparisons with \sysname.
$\mathcal{A}(Y_k)$ is pretrained with $Y_k$ which is a part of reference data $\mathbf{Y}$. An input sample $x_i$ from an unknown set $X$ is given to $\mathcal{A}(Y_k)$. $\mathcal{A}(Y_k)$ reconstructs an output sample $\hat{x}_i$. Here, we evaluate $\delta (x_i, \hat{x}_i)$, i.e., the difference between the input $x_i$ and the output $\hat{x}_i$. There could be various choices of the function $\delta$, although this paper applies the same loss function used during training $\mathcal{A}(Y_k)$ which is either MSE or iSSIM. 
$\Delta(X|Y_k)$ denotes the mean of all $\delta (x_i, \hat{x}_i)$ resulting from $\forall x_i \in X$. We conjecture that $\Delta(X|Y_k)$ would be a predictor of a similarity metric of the unknown set $X$ with respect to the reference set $Y_k$, which grows as $X$ is more dissimilar from $Y_k$. 
Furthermore, for multiple unknown sets of data $X$, $W$, ... , $Z$, we conjecture that $\Delta(X|Y_k)$, $\Delta(W|Y_k)$, ... , $\Delta(Z|Y_k)$ would predict the ordered similarity of $X$, $W$, ... , $Z$ with respect to $Y_k$. 
For example, by letting $\delta = \mathrm{MSE}$ on the results in Figure~\ref{fig:reconstructed_digits}, we obtain an ordered list $\Delta(9|4) < \Delta(7|4) < \Delta(5|4) < \Delta(3|4)$ in an increasing order of dissimilarity with respect to the set of digit 4.


We conjecture that $\Delta(X|Y)$ may reflect not only the apparent similarity at the sample space but also certain `deep criteria' based on the knowledge of $Y$ that $\mathcal{A}(Y)$ learned at pretraining. 
If unknown dataset $X$ embeds more deep features compatible with those of $Y$, we may observe smaller $\Delta(X|Y)$ from $\mathcal{A}(Y)$.
Note that, unlike task-specific supervised models, the deep features learned and extracted in \sysname are task-agnostic.
However, we acknowledge that these are yet hypothetical and not straightforward to verify. 
The results shown in Figure~\ref{fig:reconstructed_digits} suggest a possible order of digits in terms of their similarities to the digit of 4. 
Although this particular order likely concur with visually perceived similarity between digits, reasoning such an ordering may not be always obvious. 


For the rest of the paper, we demonstrate the similarity relation predicted by \sysname could be a useful indicator applicable to the typical context of transfer learning and data augmentation. We also discuss that \sysname implies runtime advantages in making an informed data-selection decision.

\begin{figure*} 
    \centering
    \includegraphics[trim=22mm 0 0mm 0, clip=true, angle=90, width=0.85\linewidth]{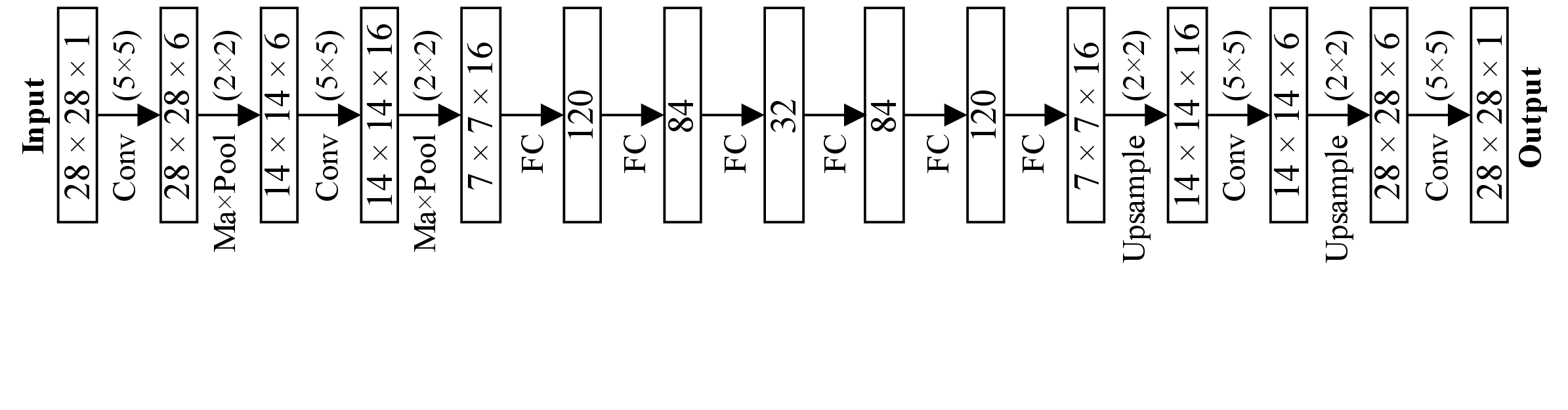}
    \caption{Layer specifications of the autoencoders used in this paper.}
    \label{fig:layer_spec}
\end{figure*}
\section{Experiments}

\begin{figure}[ht]
    \centering
    \begin{subfigure}{0.27\columnwidth}
        \centering
        \includegraphics[width=1.0\textwidth]{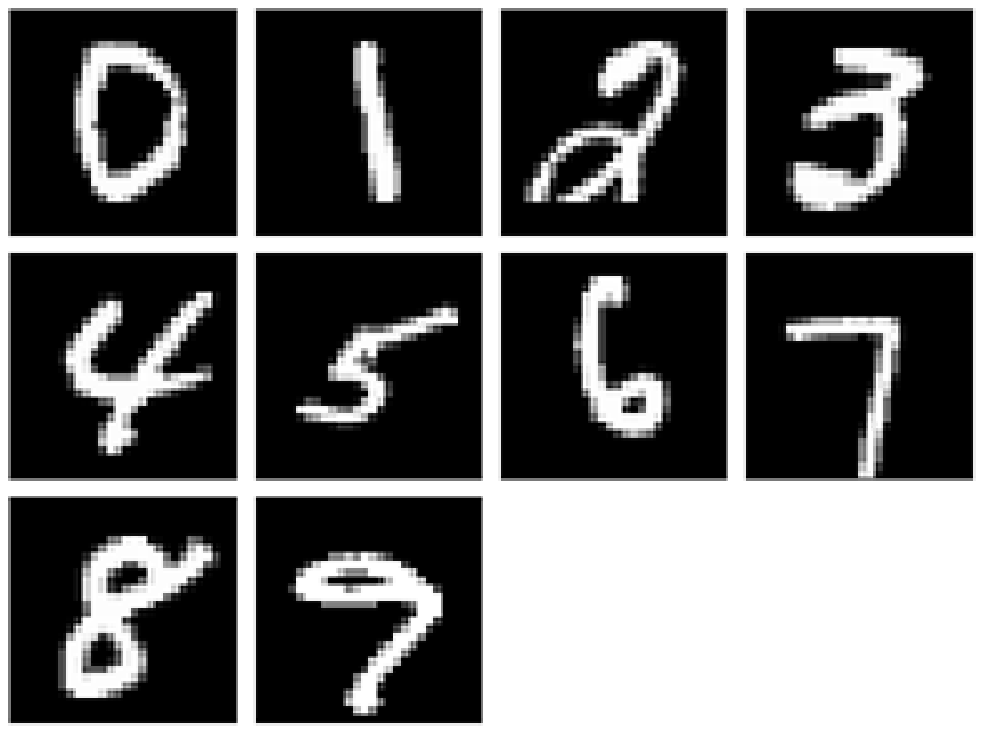}
        \caption{MNIST}
        \label{fig:example_mnist}
    \end{subfigure}
    \begin{subfigure}{0.27\columnwidth}
        \includegraphics[width=1.0\textwidth]{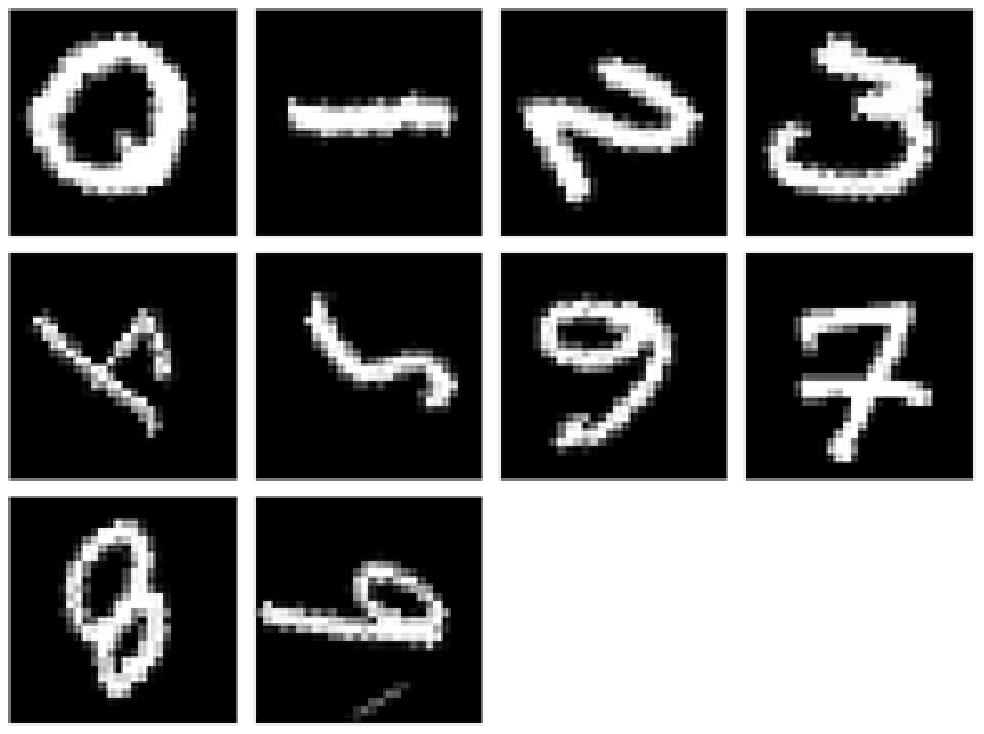}
        \caption{ROTATED}
        \label{fig:example_rotated}
    \end{subfigure}
    \begin{subfigure}{0.27\columnwidth}
        \includegraphics[width=1.0\textwidth]{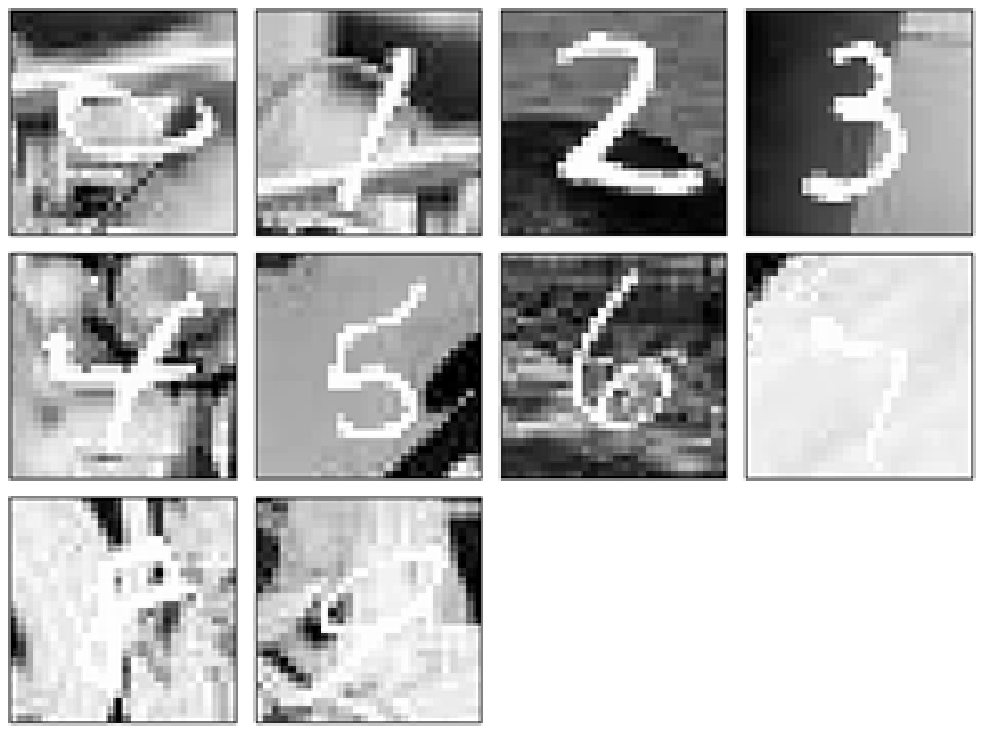}
        \caption{BGROUND}
        \label{fig:example_background}
    \end{subfigure}
    \begin{subfigure}{0.27\columnwidth}
        \centering
        \includegraphics[width=1.0\textwidth]{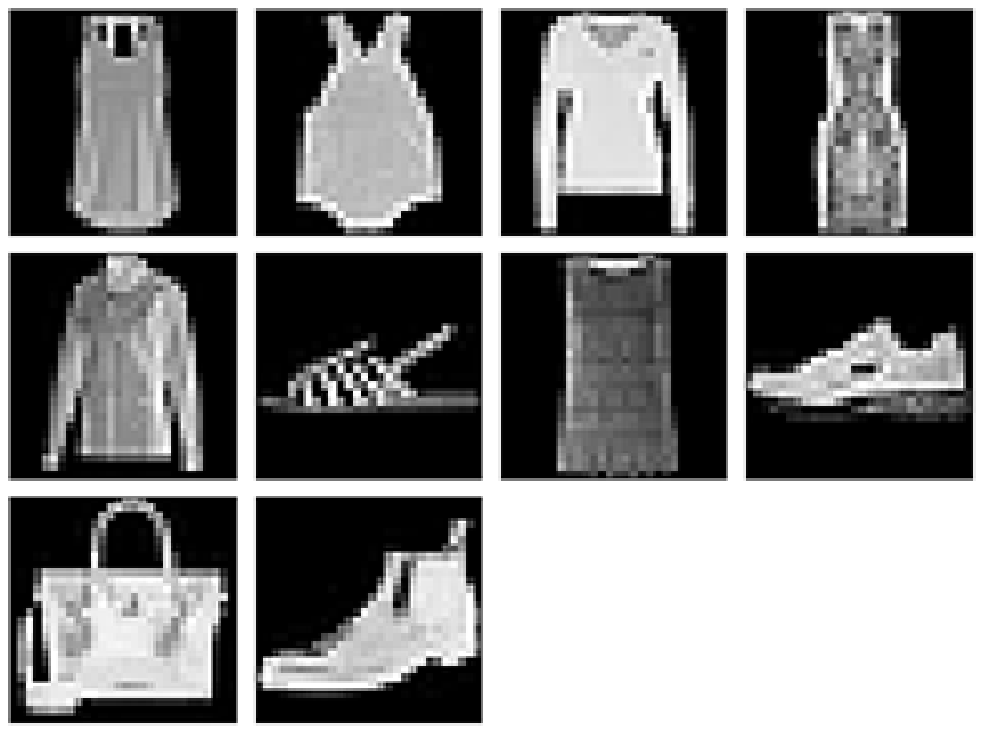}
        \caption{FASHION}
        \label{fig:example_fashion}
    \end{subfigure}
    \begin{subfigure}{0.675\columnwidth}
        \includegraphics[width=1.0\textwidth]{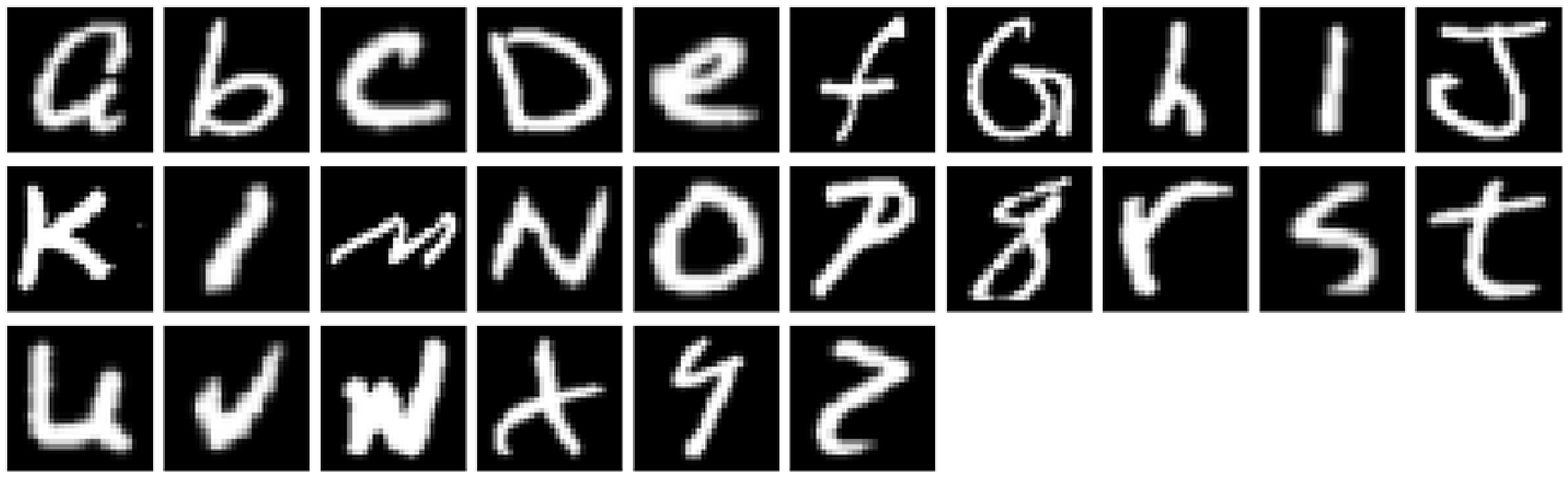}
        \caption{EMNIST}
        \label{fig:example_emnist_letters}
    \end{subfigure}
    \caption{MNIST-variant datasets used in the experiments.}
    \label{fig:mnist_variant_datasets}
\end{figure}

In this section, we explore the potential usefulness of \sysname through a series of experiments in which \sysname is applied to predict the ordered similarity relationship among datasets and among classes. 
We juxtapose our results with the similarity relationship from typical baseline methods, and find the correlations in between. 
Specifically, we experiment with five publicly available MNIST-variant datasets: MNIST~\cite{lecun1998gradient} (denoted by MNIST or $\mathbb{M}$ hereafter), rotated MNIST (ROTATED or $\mathbb{R}$ )~\cite{larochelle2007empirical}, background-image MNIST (BGROUND or $\mathbb{B}$ )~\cite{larochelle2007empirical}, fashion MNIST (FASHION or $\mathbb{F}$ )~\cite{xiao2017fashion}, and EMNIST-Letters (EMNIST or $\mathbb{E}$ )~\cite{cohen2017emnist}. 
$\mathbb{E}$ consists of 26 classes of English alphabet, while other datasets have 10 classes each. 
Figure~\ref{fig:mnist_variant_datasets} illustrates the datasets. 

The autoencoders used in this paper have a symmetric architecture whose encoder part is largely adopted from LeNet-5~\cite{lecun1998gradient}, while \sysname is open to other autoencoder models.
Figure~\ref{fig:layer_spec} lists the layer specifications. 
For loss functions, we used the mean squared error (denoted MSE hereafter) or the inverted structural similarity metric (denoted iSSIM hereafter, i.e., $\mathrm{iSSIM} = 1-\mathrm{SSIM}$)~\cite{wang2004image}, while \sysname is open to other choices of loss function. 
We conducted our experiments on machines with one Intel i7-6900K CPU at 3.20 GHz and four NVIDIA GTX 1080 Ti GPUs.

\subsection{Predicting inter-dataset similarity} \label{sec:predicting_inter_dataset}

\begin{figure*}[ht]
    \centering
    \begin{subfigure}{0.310\textwidth}
        \centering
        \includegraphics[trim=0 0 12mm 10mm, clip=true, width=1.0\textwidth]{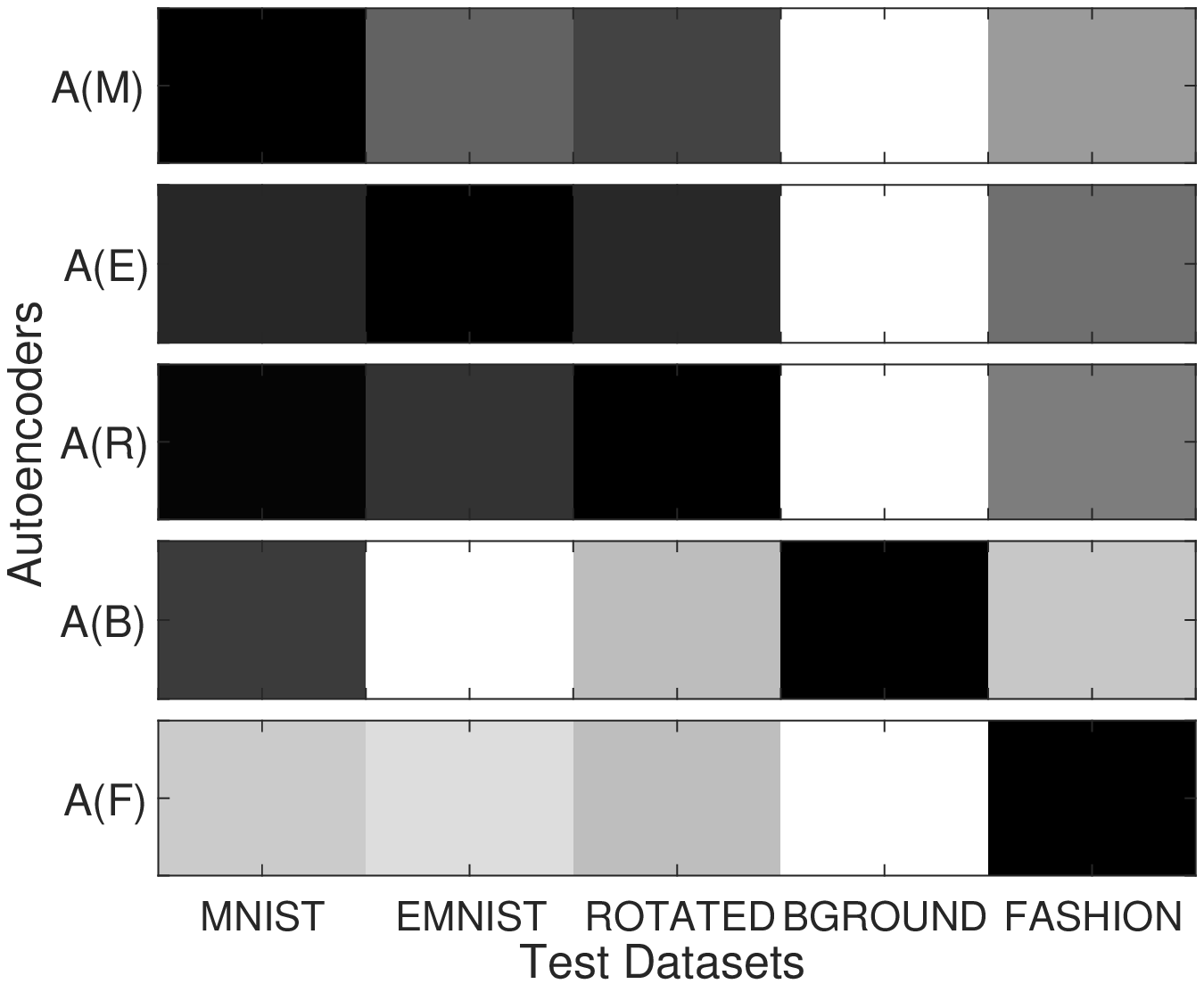}
        \caption{MSE losses from \sysname}
        \label{fig:family_AE_mse_map}
    \end{subfigure}
    \begin{subfigure}{0.310\textwidth}
        \centering
        \includegraphics[trim=0 0 12mm 10mm, clip=true, width=1.0\textwidth]{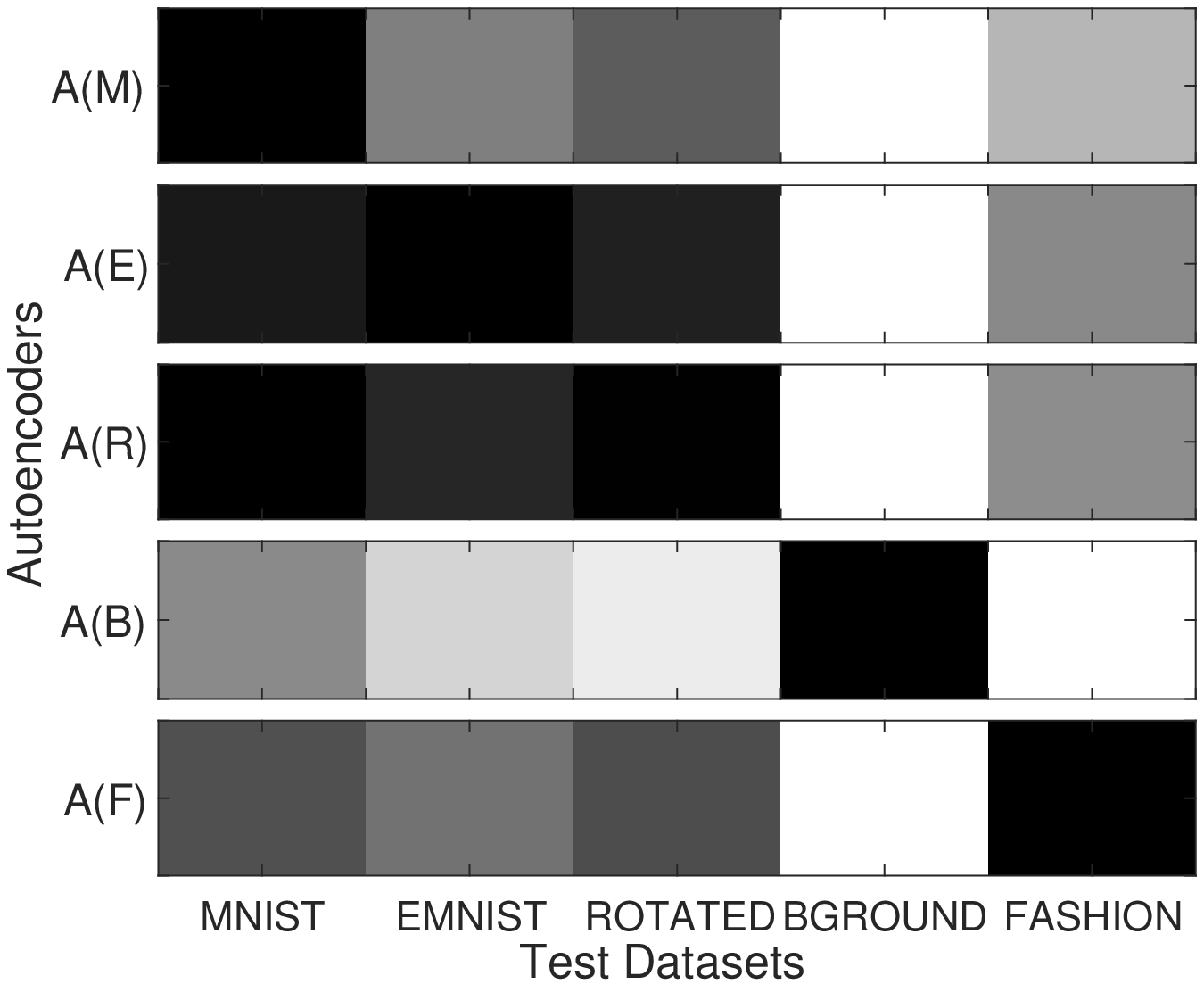}
        \caption{iSSIM losses from \sysname}
        \label{fig:family_AE_dssim_map}
    \end{subfigure}
    \begin{subfigure}{0.350\textwidth}
        \centering
        \includegraphics[trim=0 0 12mm 10mm, clip=true, width=1.0\textwidth]{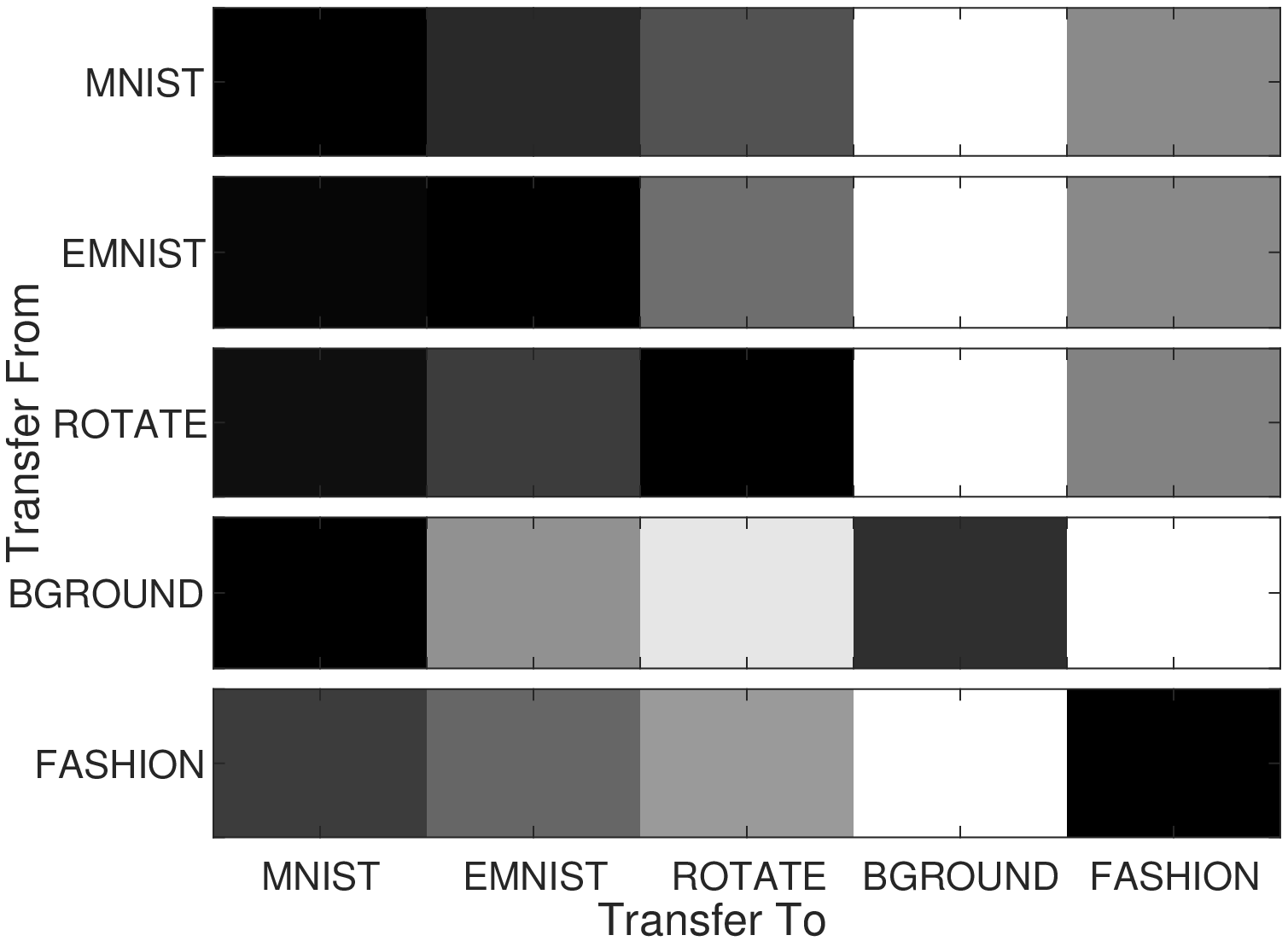}
        \caption{Accuracy of retrained networks}
        \label{fig:family_model_transfer_map}    
    \end{subfigure}
    \caption{Inter-dataset similarity. Darker means: (a), (b) smaller losses; (c) higher accuracy.}
    \label{fig:inter_dataset_similarity}
\end{figure*}

In this experiment, we demonstrate the quality of the \sysname-predicted inter-dataset similarities as explained in Section~\ref{sec:scenarios}.
We train 5 autoencoders, one for each of our datasets. 
Then each autoencoder is given the samples from the other 4 datasets, and the resulting $\Delta(X|Y)$ is computed. 
We conducted the same experiments with MSE and iSSIM losses for \sysname. 
Figure~\ref{fig:family_AE_mse_map} and \ref{fig:family_AE_dssim_map} depict the relative $\Delta(X|Y)$ levels from \sysname with MSE and iSSIM loss functions, respectively.  

For baseline, we trained five 10-class classifiers using LeNet-5 model, one for each dataset. 
For EMNIST, we used only the first 10 classes (A through J). 
After each base network is trained, 
we froze all the convolution layers of the base networks and retrained each network's FC layers with the four other datasets, resulting in (5 base networks) $\times$ (4 different datasets per base dataset) $=$ 20 retrained networks. 
Figure~\ref{fig:family_model_transfer_map} depicts the accuracy of retrained networks normalized by the original accuracy of the base network. 
We balanced the number of samples across classes and across datasets. Each dataset was divided to 5:1 for training and testing. We normalized the color charts for visibility, as only the relative differences matter below.

\begin{table}[ht]
    \centering
    \caption{Spearman's $\rho$ between \sysname-predicted orders and retrained accuracy orders}
    \label{tab:spearman_correlation}
    \begin{small}
    \begin{tabular}{cccccc}
        \toprule
        Base Dataset                       & $\mathbb{M}$   & $\mathbb{E}$  & $\mathbb{R}$ & $\mathbb{B}$ & $\mathbb{F}$ \\ 
        \midrule
        $\rho$ (for AEs w/ MSE)       & 0.9     & 1.0     & 1.0     & 0.6     & 0.7     \\
        $\rho$ (for AEs w/ iSSIM)     & 0.9     & 1.0     & 0.9     & 0.9     & 0.9     \\ 
        \bottomrule       
    \end{tabular}
    \end{small}
\end{table}

To assess the consistency between the \sysname-predicted similarity relationship and the baseline of retraining-implied similarity relationship, we compare the ordered lists of datasets $X_i \in \{\mathbb{M}$, $\mathbb{E}$, $\mathbb{R}$, $\mathbb{B}$, $\mathbb{F}\}$ from both methods: one ordered by the test losses $D(X_i|Y)$ from each $\mathcal{A}(Y)$ and the other ordered by the retrained accuracy transferred from each base network (denoted by $\mathcal{B}(Y)$), where $Y$ denotes base dataset that $\mathcal{A}(Y)$ and $\mathcal{B}(Y)$ were previously trained with.
For example, $\mathcal{A}(\mathbb{M})$ with MSE  predicts the loss-ordered list of $(\mathbb{M}$, $\mathbb{R}$, $\mathbb{E}$, $\mathbb{F}$, $\mathbb{B})$ (shown in Figure~\ref{fig:family_AE_mse_map}), while retraining $\mathcal{B}(\mathbb{M})$ gives the accuracy-ordered list of $(\mathbb{M}$, $\mathbb{E}$, $\mathbb{R}$, $\mathbb{F}$, $\mathbb{B})$ as shown in Figure~\ref{fig:family_model_transfer_map}. 

Table~\ref{tab:spearman_correlation} lists Spearman's rank correlation coefficients ($\rho$) from each pair of \sysname-predicted and retrain-implied similarity orderings, both sharing the same base dataset. Spearman's $\rho$ is a popular measure of the correlation between ordinal variables, ranging between [-1, 1] where 1 means a pair of identically ordered lists and -1 means fully reversely ordered. The table shows reasonably high correlations. The results indicate that using the iSSIM loss function in \sysname yields better correlations, possibly due to the higher robustness in taking account of the structural similarity. 

This experiment presents a supportive example that \sysname-predicted similarity relationship would be reasonably consistent with those implied by conventional transfer learning practices. If consistent indeed, then what would be the advantage of \sysname? It is the runtime efficiency, as predicting the similarity relationship by using \sysname does not require any training at comparison time. We demonstrate the latency experiment results in the following subsection.

\subsection{Latency of predicting inter-dataset similarity}
In many real applications, estimating the similarity between data could be a practical issue in terms of computational complexity and the latency involved therein. For example, popular service platforms should deal with massive influx of new data; a mid-2013 article reports that more than 350 million photos being uploaded to Facebook every day\footnote{\url{https://www.businessinsider.com/facebook-350-million-photos-each-day-2013-9}}, and the CEO of YouTube revealed that 400 hours of content was being uploaded to YouTube every minute in mid-2015\footnote{\url{https://www.tubefilter.com/2015/07/26/youtube-400-hours-content-every-minute/}}. Upon incoming arbitrary data, predicting the characteristics of the new data with respect to the reference data or models that various service APIs rely on could be an early step that occurs frequently. Furthermore, interactive or real-time services are highly latency-sensitive, such as product identification by mobile computer vision\footnote{\url{https://www.amazon.com/b?ie=UTF8&node=17387598011}}. 

In this experiment, we demonstrate the latency measurements in the context of the previous experiments presented in Section~\ref{sec:predicting_inter_dataset}, i.e., predicting the inter-dataset similarity by \sysname and the baseline methods based on transfer learning. In both approaches, there are two types of latency: (1) the one-time latency to build the pretrained models for each reference dataset, i.e., scratch-training the autoencoders (\sysname) and the classifiers (baselines), and (2) the runtime latency to compare an arbitrary dataset against the reference datasets, i.e., inferring with the autoencoders (\sysname) and transferring from the pretrained classifiers (baseline). As the latter takes place whenever new incoming data is to be compared against the reference datasets, we conjecture that the runtime latency would matter much more in many services with large-scale or interactive requirements. Below we discuss the latency results for the latter, followed by the results for the former. 

\begin{figure}[ht]
    \centering
    \begin{subfigure}{1.0\columnwidth}
        \includegraphics[width=1.0\textwidth]{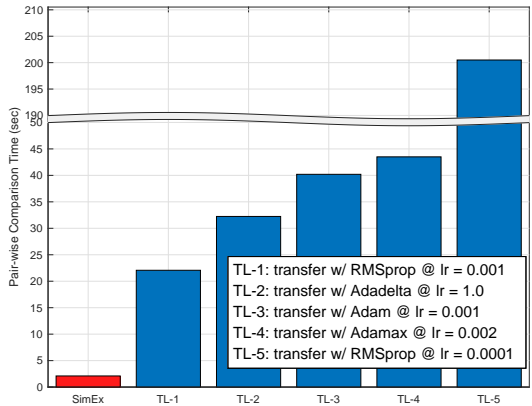}
    \end{subfigure}
    \caption{Latencies for pair-wise similarity prediction.}
    \label{fig:timing_comparison_times}
\end{figure}

Figure~\ref{fig:timing_comparison_times} depicts the mean runtime latency values averaged from 5 measurements per configuration. The error bars are omitted for brevity as the training times are highly consistent and thereby the standard deviations are insignificantly small, e.g., around 2\% of the mean. The configurations include \sysname and 5 different baseline configurations denoted by TL-1 through TL-5 with varying optimizers and learning rates. (TL stands for `transfer learning'.) The latency values reported here are for pair-wise comparison, i.e., comparing the unknown dataset against one reference dataset. For \sysname, the latency indicates the total time to have all samples in the new dataset forward-pass an autoencoder pretrained with respect to a reference dataset. For baselines, the latencies indicate the time elapsed until the transfer learning hits the minimum loss. All measurements were done on the identical hardware and framework setup. 

\sysname features 2.103 seconds per inference which is more than 10 times faster than the best performing baseline at TL-1 configuration that completes within 22.07 seconds / 8 epochs with RMSprop optimizer and its default learning rate ($1.0\times10^{-3}$) to transfer the reference classifier with respect to the unknown dataset. Other baseline configurations of TL-2 through TL-4 that use different optimizers and their default learning rates exhibit worse transfer latencies, i.e., 32.24 -- 40.20 seconds at 12 -- 19 epochs, respectively. In fact, it would be an aggressive strategy to use the default learning rates for transfer learning; it is often a common practice to fine-tune the transferring model with a smaller learning rate. TL-5 reflects such a case in that the learning rate is reduced to 10\% of the default value used in TL-1. Not surprisingly, TL-5 latency is roughly 10 times longer than TL-1. Further increasing the learning rates beyond the default values might accelerate the runtime latencies of the baselines, but we observed degrading classification accuracy in exchange. 

Now that we verified \sysname's runtime latency advantage outperforming the baselines by more than an order of magnitude, we examine the one-time latency to build the pretrained models for each reference dataset. 

We found a tricky issue with training the \sysname autoencoders that the loss slowly ever-decreases for a very long time, taking 5+ hours / 4000+ epochs to reach the minimum loss. In contrast, for the baselines, scratch-training the base classifiers converges at their minimum losses a lot more quickly, e.g., taking 23.49 -- 33.45 seconds / 7 -- 14 epochs. Even though training for the reference datasets takes place only one time per reference dataset and it could be done offline before services are active, \sysname's one-time latency overheads seem unarguably too large to be tolerated. 

To circumvent, we note that \sysname does not have to train its autoencoders all the way to their best, because \sysname does not pursue high-quality reconstruction; what matters is only the \textit{relative} difference between reconstructed results. Therefore, we can safely stop the training earlier at the point where the relative similarity orderings converge, and it may be reachable not necessarily with best-trained autoencoders yielding high-quality reconstruction. 

\begin{figure}[ht]
    \centering
    \begin{subfigure}{1.0\columnwidth}
        \includegraphics[trim=0 4mm 0 9mm, clip=true, width=1.0\textwidth]{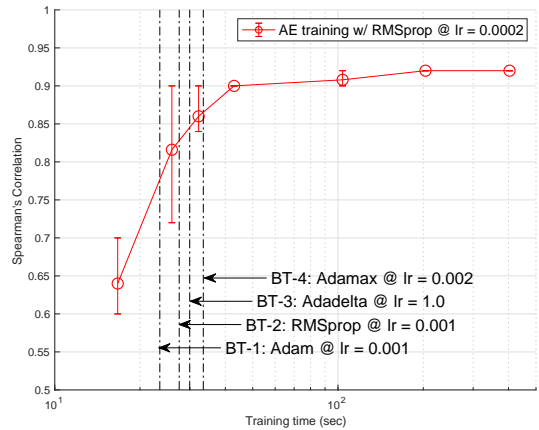}
        \caption{AE training w/ RMSprop(lr = 2.0e-4)}
        \label{fig:timing_training_vs_corr_rms0.0002}
    \end{subfigure}
    \begin{subfigure}{1.0\columnwidth}
        \includegraphics[trim=0 4mm 0 9mm, clip=true, width=1.0\textwidth]{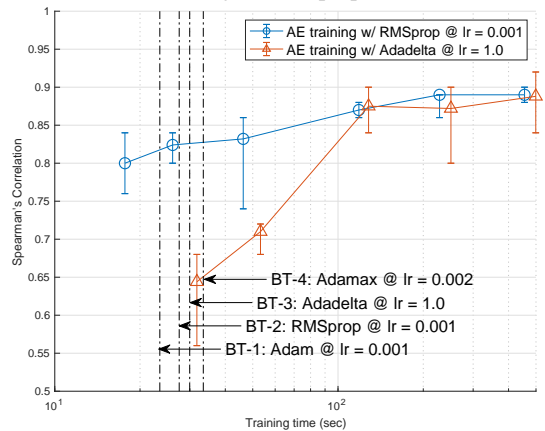}
        \caption{AE training w/ RMSprop(lr = 1.0e-3) and Adadelta(lr = 1.0)}
        \label{fig:timing_training_vs_corr_rms0.001_adadelta1.0}
    \end{subfigure}
    \caption{Interim training time vs. Spearman's correlation}
    \label{fig:timing_training_vs_corr}
\end{figure}

We examined this strategy. In Figure~\ref{fig:timing_training_vs_corr_rms0.0002}, the red solid line depicts the results for \sysname trained with RMSprop optimizer and lr = $2.0\times10^{-4}$ along varying training epochs of 3, 5, 7, 10, 25, 50, and 100. The x-axis represents the time elapsed until each epoch, and the error bars represent the min and max values from 5 repeated measurements at each point. The vertical lines labeled BT-1 through BT-4 represent the one-time latencies for the baseline methods to train their base classifiers (BT stands for `base training'). 

At each training epoch, we compared the similarity orderings inferred by the interim autoencoder model at that epoch against the similarity orderings from the baseline method, which is represented by Spearman's correlation coefficients ($\rho$) in Figure~\ref{fig:timing_training_vs_corr_rms0.0002}. We found that, at 203.6 sec / 50 epochs, the interim autoencoder's $\rho$ reaches the final model's $\rho=0.92$ with zero standard deviation across 5 repeated measurements, and remains the same for further epochs. If we tolerate a slight difference, the interim autoencoder achieves a stable, non-fluctuating $\rho=0.90$ at 42.95 sec / 10 epochs, which is slightly less than the twice of the fastest-training baseline classifier. 

Note that we used a reduced learning rate in Figure~\ref{fig:timing_training_vs_corr_rms0.0002} below the default value of RMSprop ($1.0\times10^{-3}$). We have experimented various combination of optimizers and learning rates, from which the configuration in Figure~\ref{fig:timing_training_vs_corr_rms0.0002} exhibit the quickest and highest convergence towards the final $\rho$. Counterintuitively, we found a general trend that a learning rate smaller than the default rate per optimizer actually helps the model reach the final $\rho$ more quickly. For comparison, Figure~\ref{fig:timing_training_vs_corr_rms0.001_adadelta1.0} illustrates the configuration with RMSprop and its default learning rate of $1.0\times10^{-3}$, along with another configuration out of many we have experimented. We leave further investigation on this counterintuitive trend to the future work. 

The results presented in this experiment were collected under iSSIM loss across all measurements. We observed very similar trends under MSE loss, thereby omitting the results.

\begin{figure*}[ht]
    \centering
    \begin{subfigure}{0.245\textwidth}
        \centering
        \includegraphics[trim=0 0 10mm 0mm, clip=true, width=1.0\textwidth]{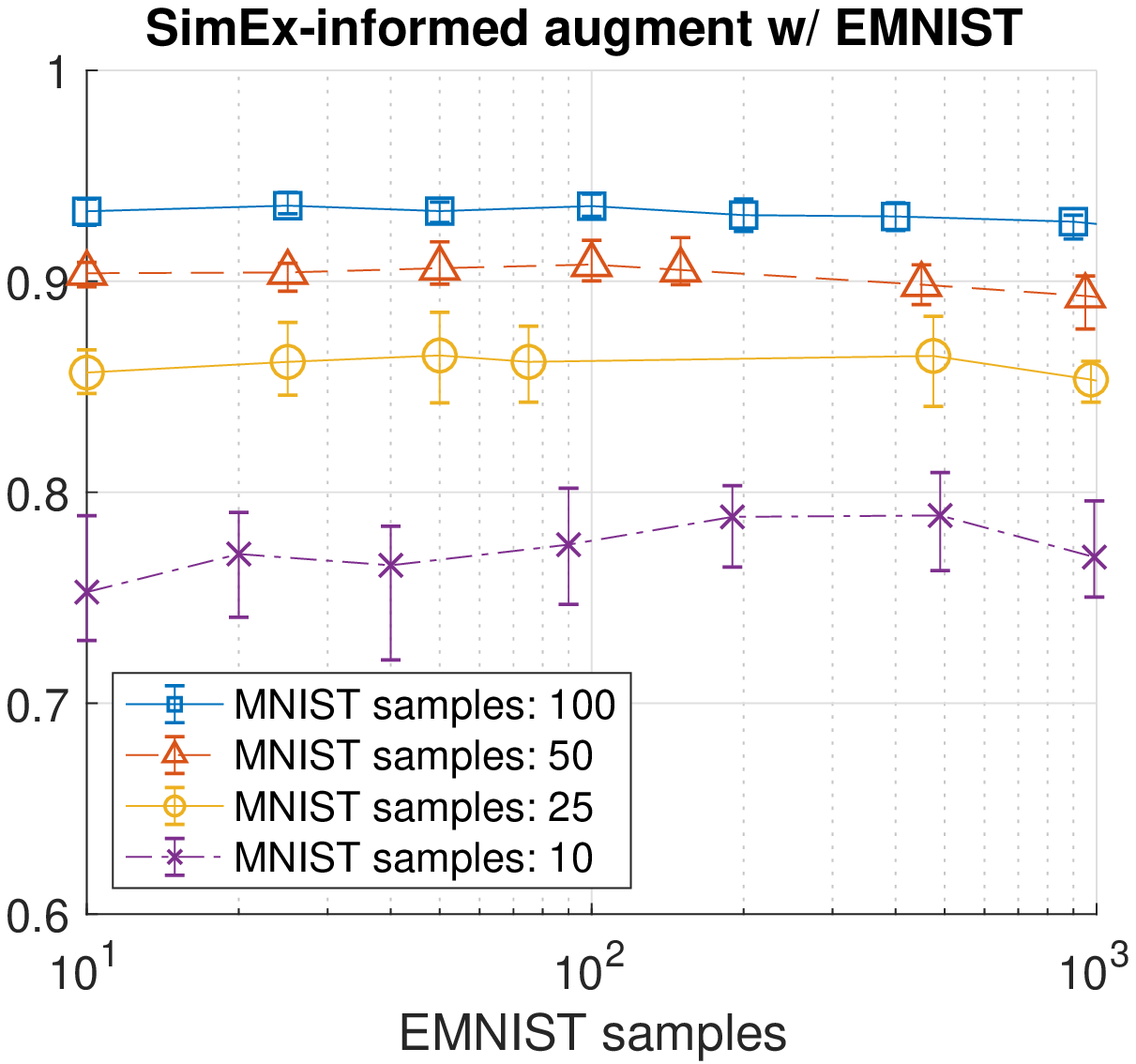}
        \label{fig:augmentation_ae_mse_emnist}
    \end{subfigure}
    \begin{subfigure}{0.245\textwidth}
        \centering
        \includegraphics[trim=0 0 10mm 0mm, clip=true, width=1.0\textwidth]{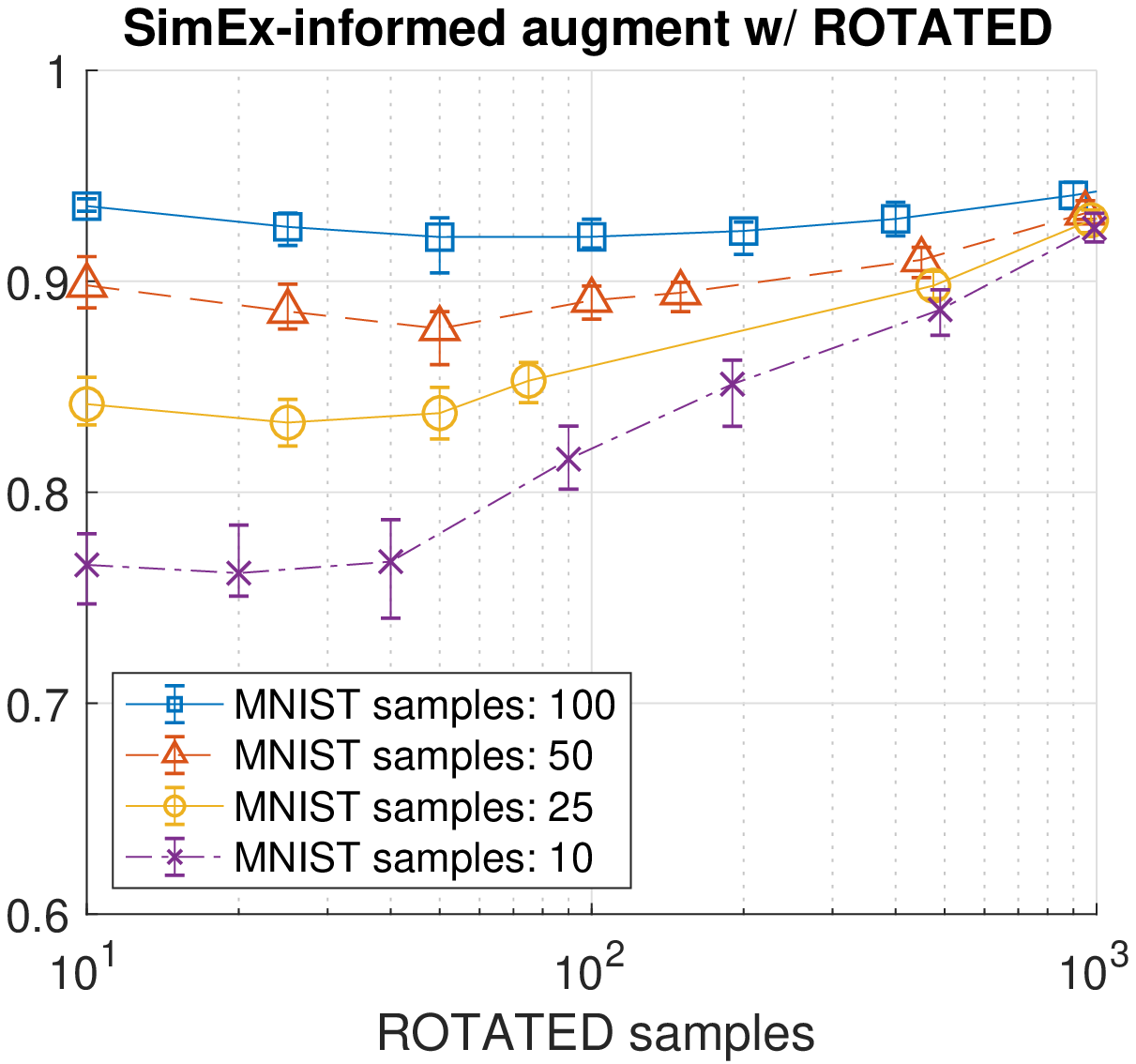}
        \label{fig:augmentation_ae_mse_rotated}
    \end{subfigure}
    \begin{subfigure}{0.245\textwidth}
        \centering
        \includegraphics[trim=0 0 10mm 0mm, clip=true, width=1.0\textwidth]{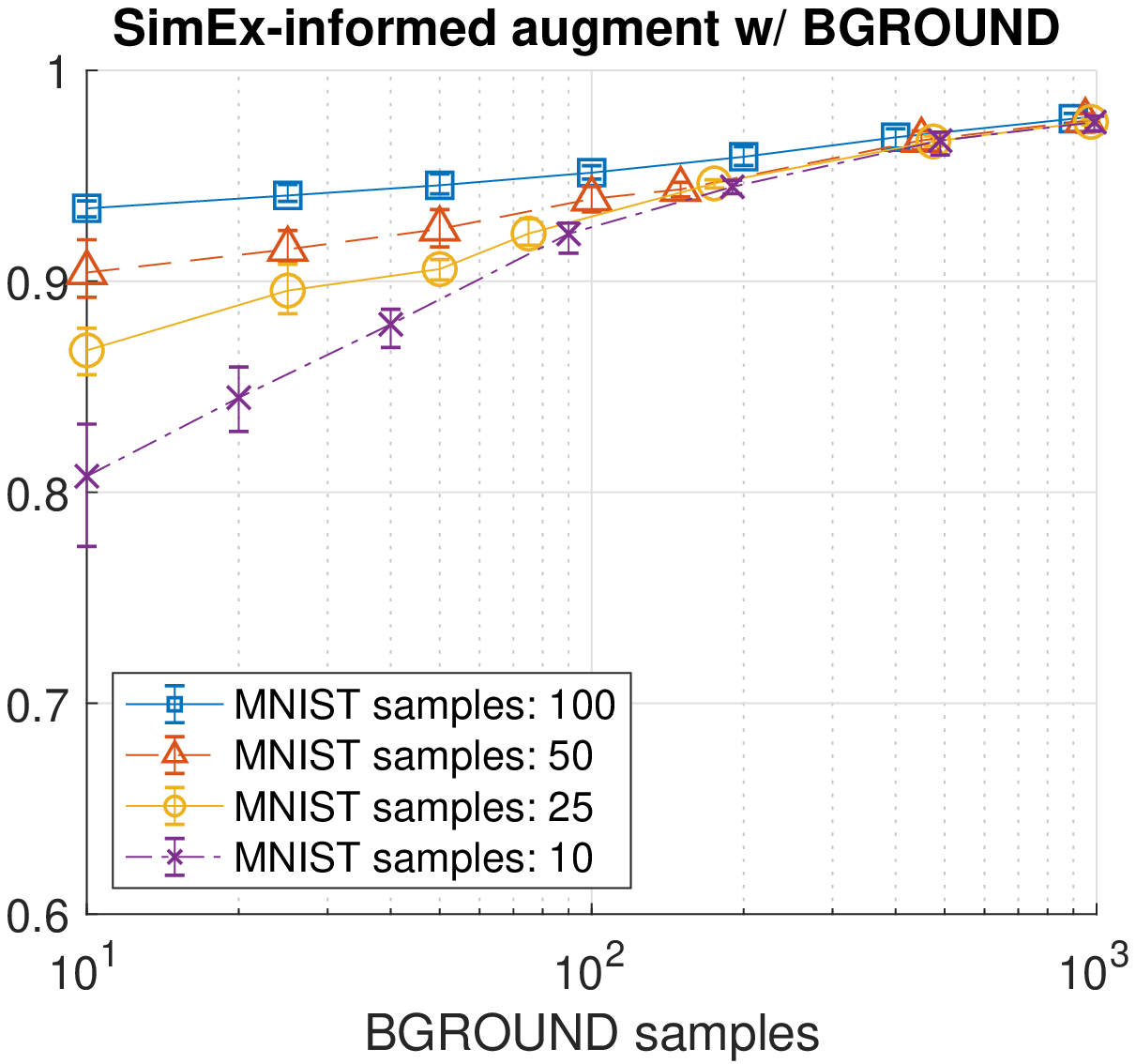}
        \label{fig:augmentation_ae_mse_bground}    
    \end{subfigure}
    \begin{subfigure}{0.245\textwidth}
        \centering
        \includegraphics[trim=0 0 10mm 0mm, clip=true, width=1.0\textwidth]{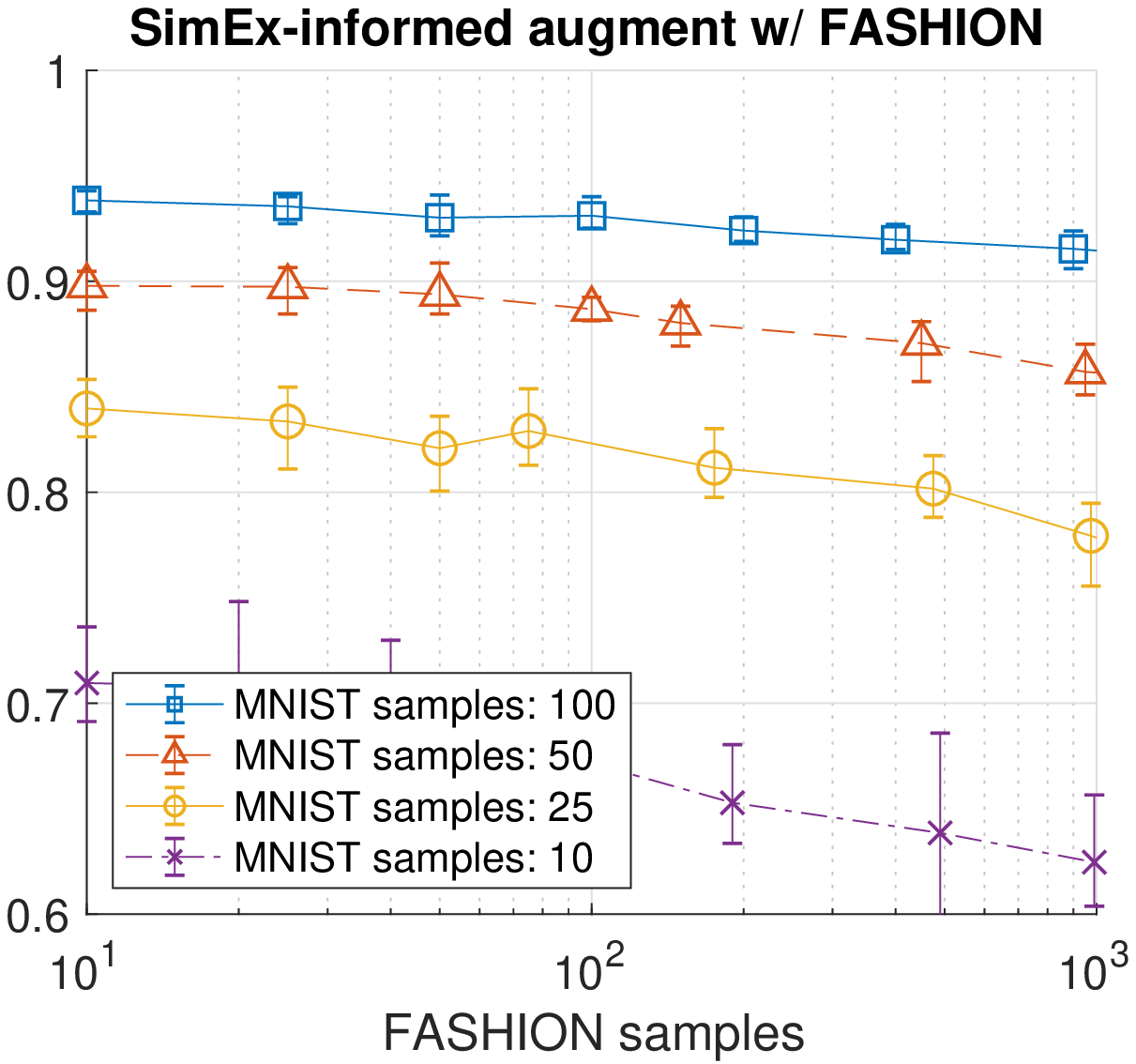}
        \label{fig:augmentation_ae_mse_fashion}    
    \end{subfigure}
    \begin{subfigure}{0.245\textwidth}
        \centering
        \includegraphics[trim=0 0 10mm 0mm, clip=true, width=1.0\textwidth]{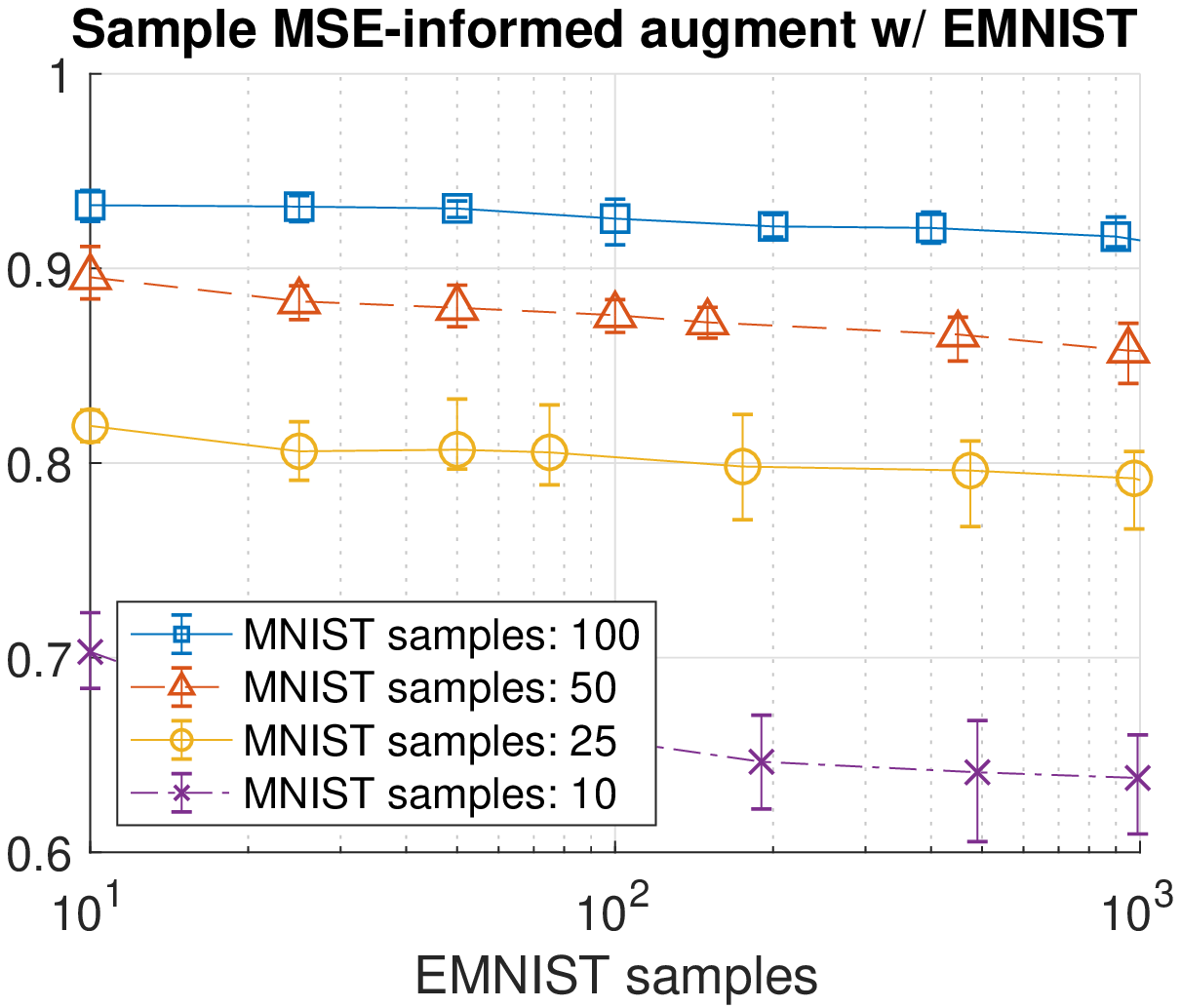}
        \label{fig:augmentation_sample_mse_emnist}
    \end{subfigure}
    \begin{subfigure}{0.245\textwidth}
        \centering
        \includegraphics[trim=0 0 10mm 0mm, clip=true, width=1.0\textwidth]{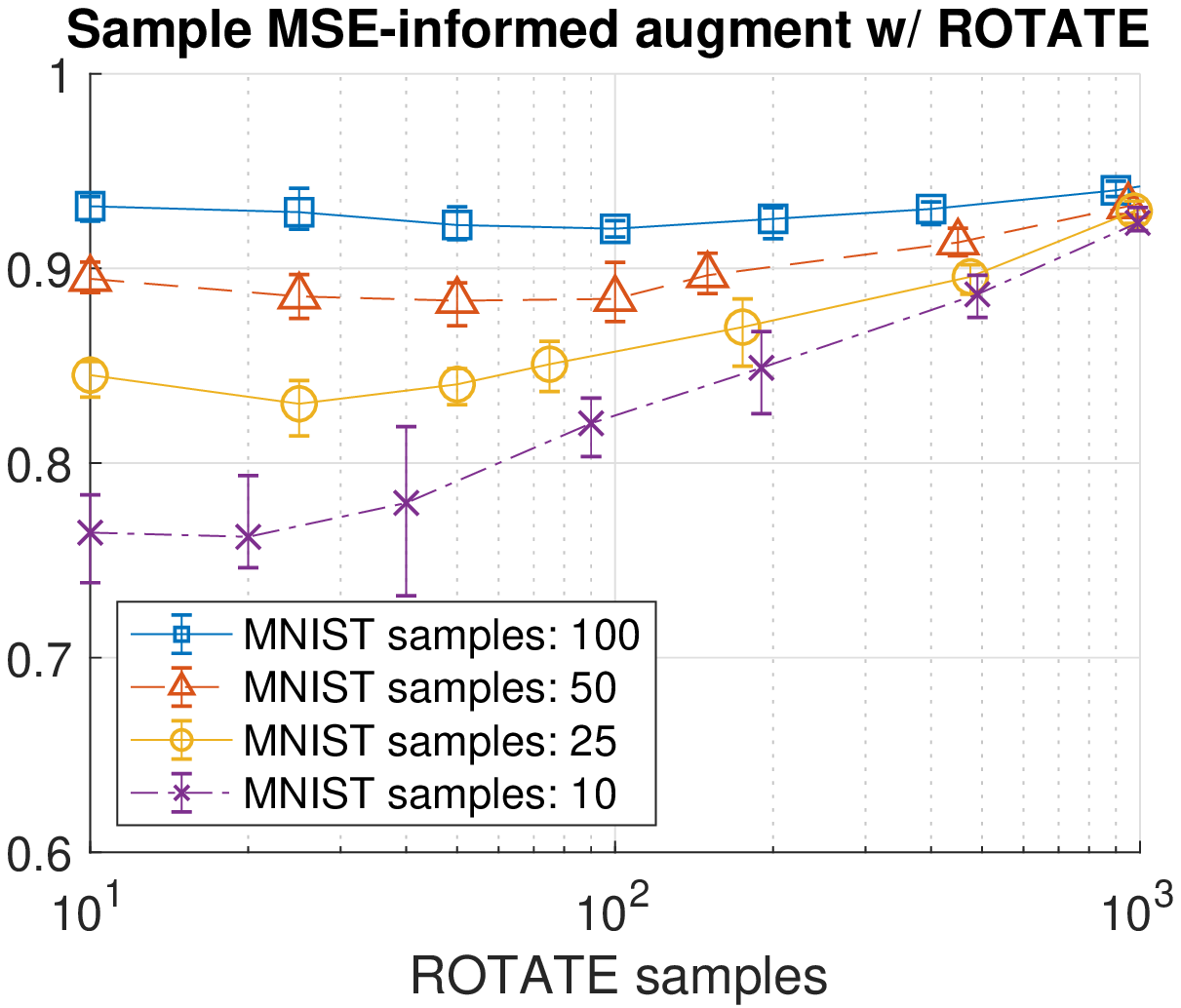}
        \label{fig:augmentation_sample_mse_rotated}
    \end{subfigure}
    \begin{subfigure}{0.245\textwidth}
        \centering
        \includegraphics[trim=0 0 10mm 0mm, clip=true, width=1.0\textwidth]{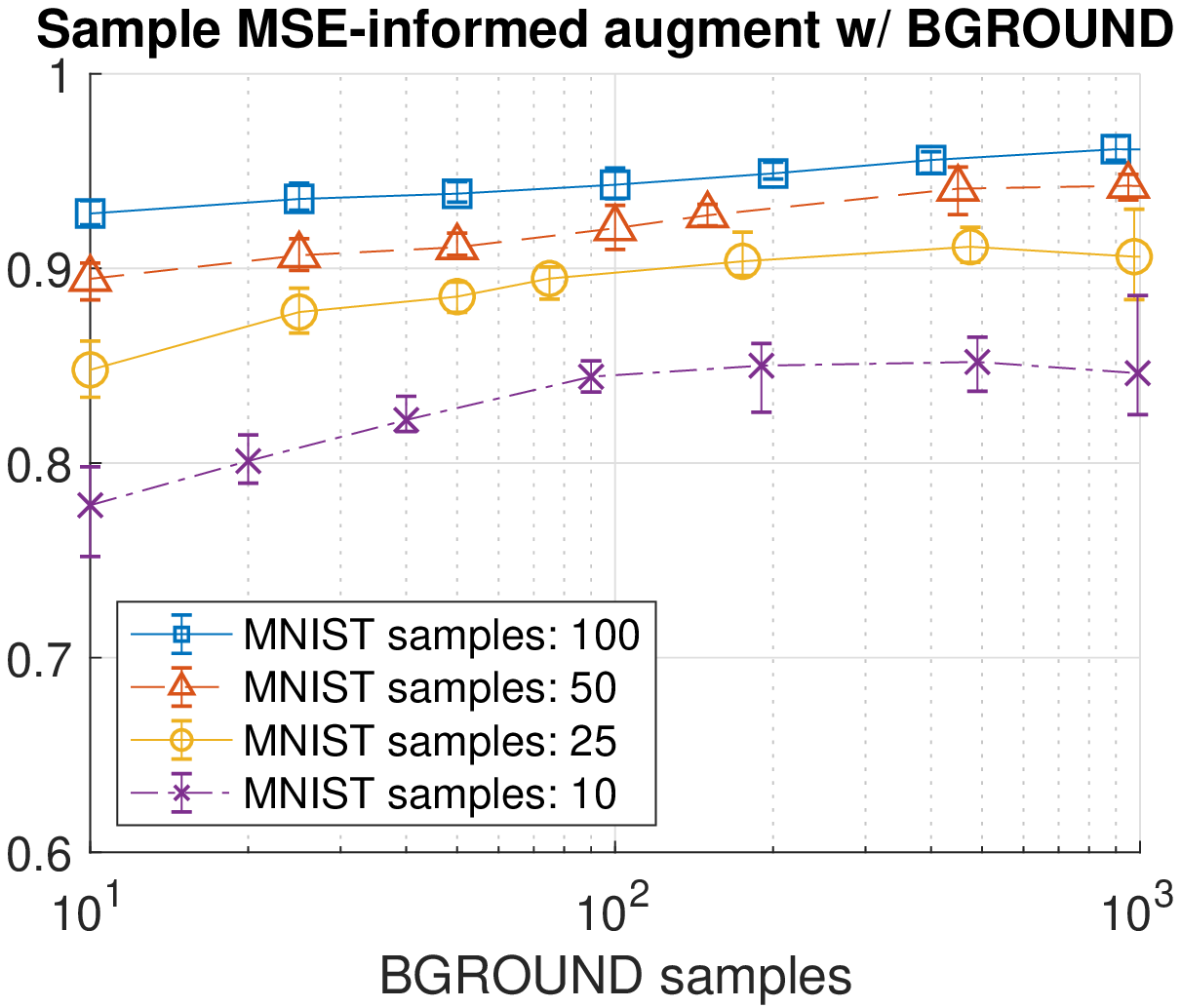}
        \label{fig:augmentation_sample_mse_bground}    
    \end{subfigure}
    \begin{subfigure}{0.245\textwidth}
        \centering
        \includegraphics[trim=0 0 10mm 0mm, clip=true, width=1.0\textwidth]{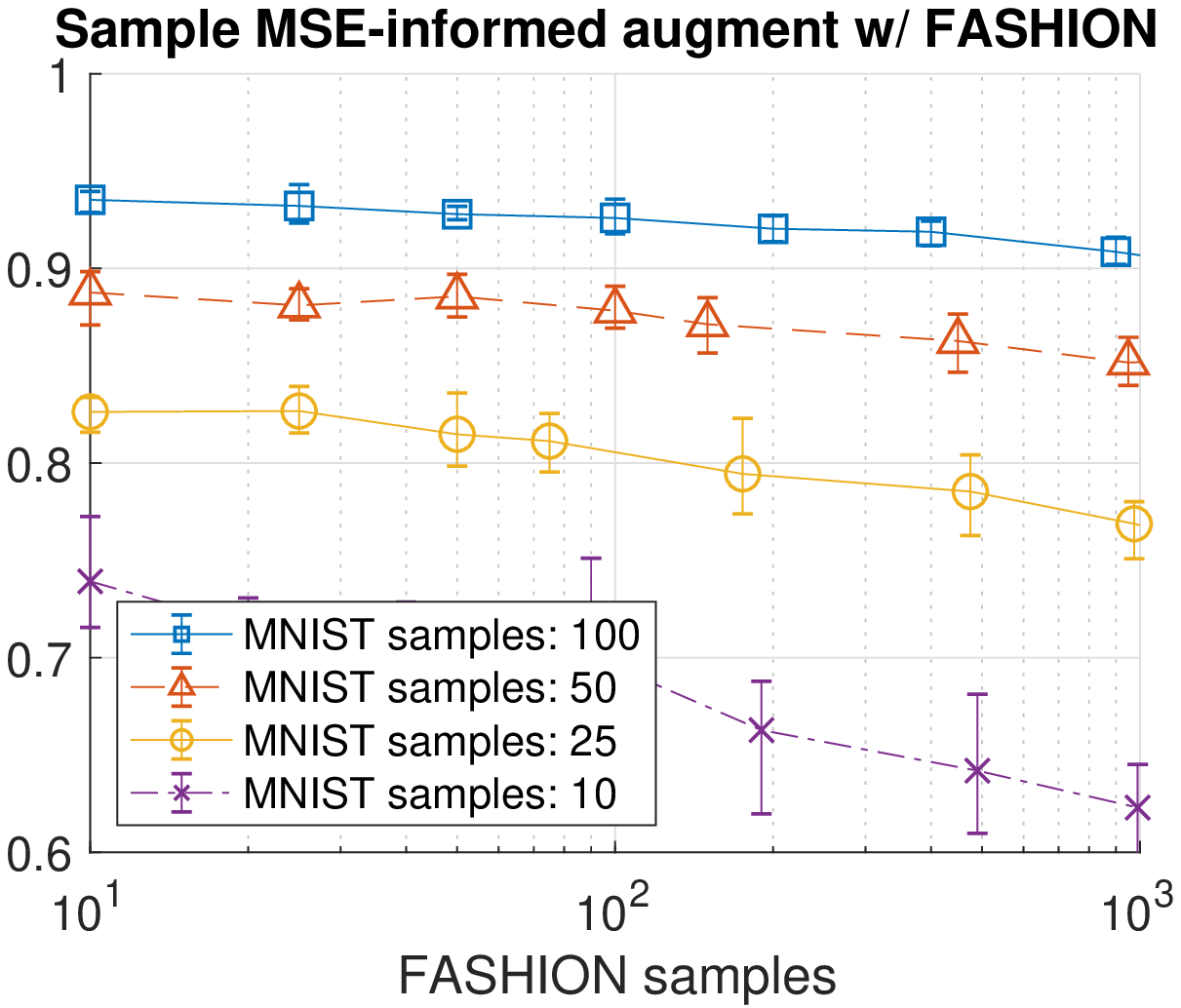}
        \label{fig:augmentation_sample_mse_fashion}    
    \end{subfigure}
    \begin{subfigure}{0.245\textwidth}
        \centering
        \includegraphics[trim=0 0 10mm 0mm, clip=true, width=1.0\textwidth]{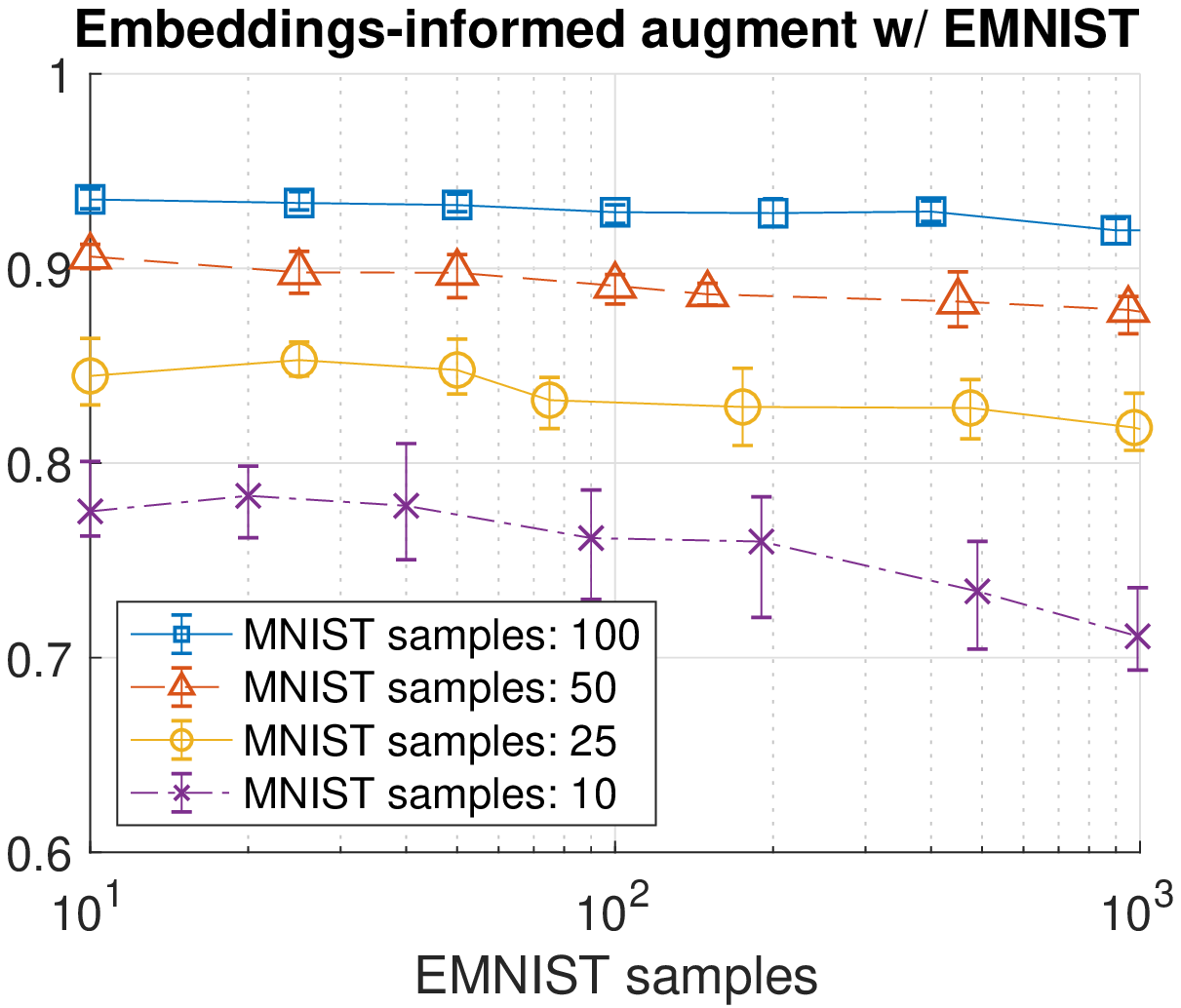}
        \label{fig:augmentation_embed_mse_emnist}
    \end{subfigure}
    \begin{subfigure}{0.245\textwidth}
        \centering
        \includegraphics[trim=0 0 10mm 0mm, clip=true, width=1.0\textwidth]{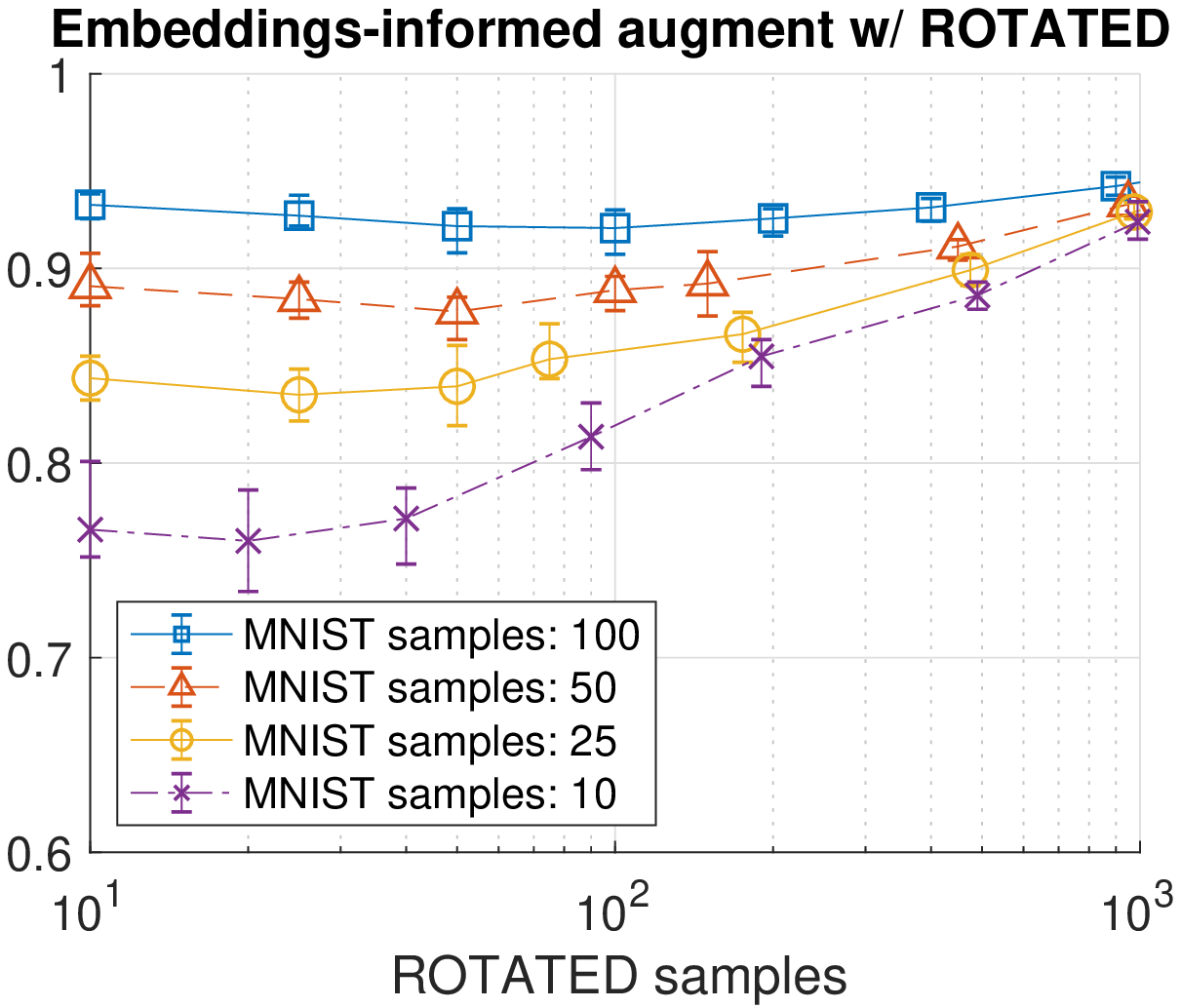}
        \label{fig:augmentation_embed_mse_rotated}
    \end{subfigure}
    \begin{subfigure}{0.245\textwidth}
        \centering
        \includegraphics[trim=0 0 10mm 0mm, clip=true, width=1.0\textwidth]{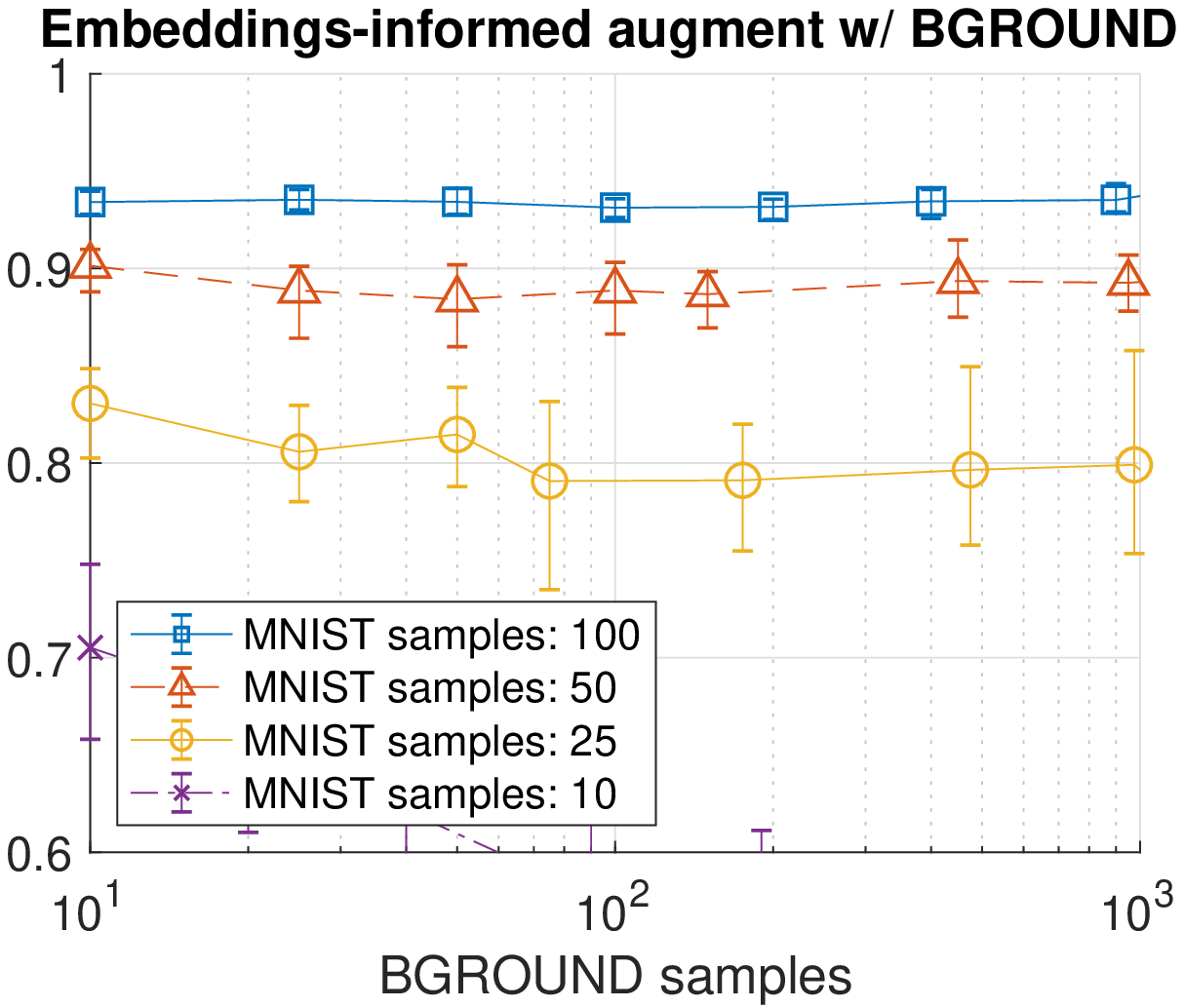}
        \label{fig:augmentation_embed_mse_bground}    
    \end{subfigure}
    \begin{subfigure}{0.245\textwidth}
        \centering
        \includegraphics[trim=0 0 10mm 0mm, clip=true, width=1.0\textwidth]{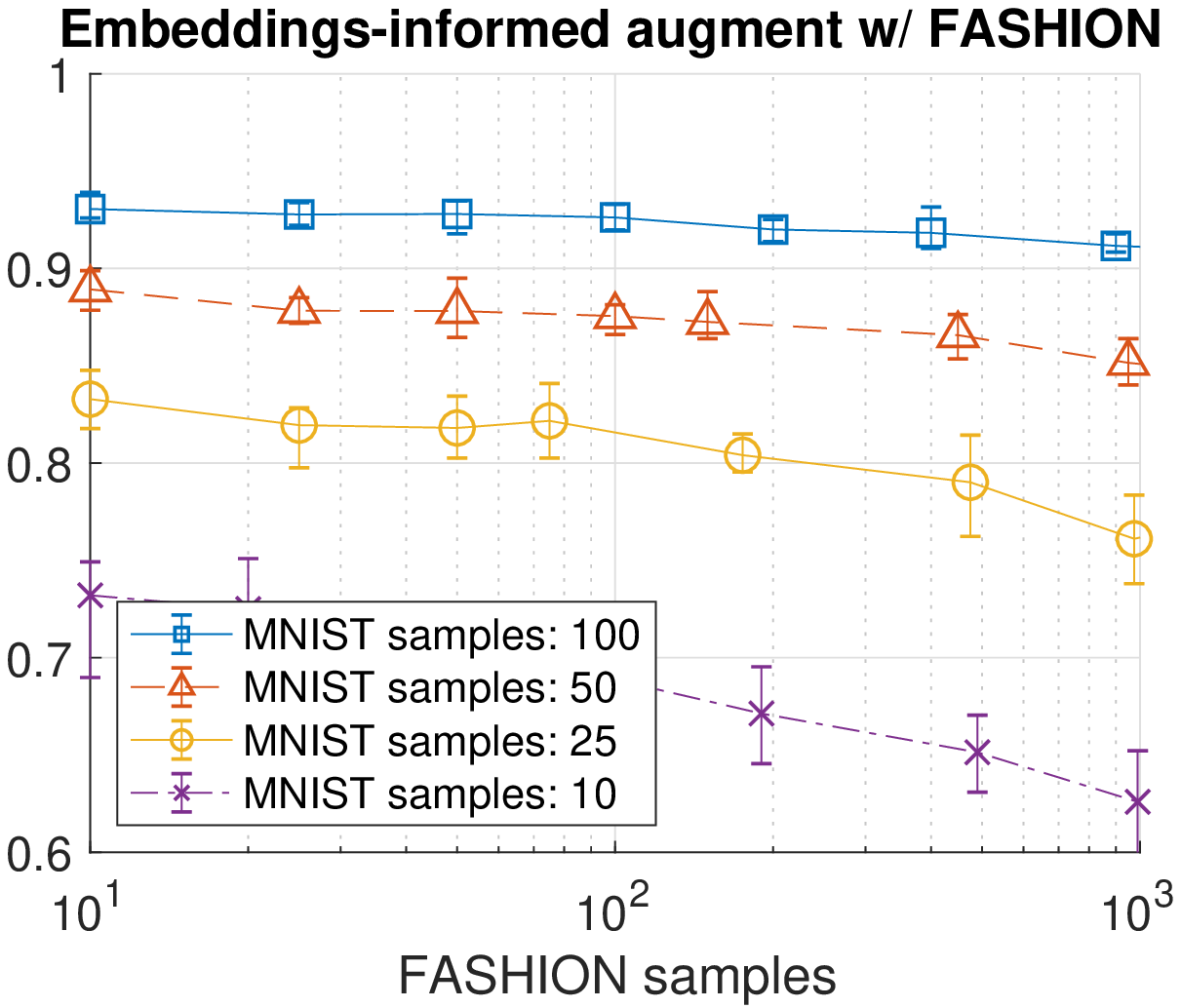}
        \label{fig:augmentation_embed_mse_fashion}    
    \end{subfigure}
    \caption{Classification accuracy after supplementing training samples from a heterogeneous dataset.}
    \label{fig:hetero_dataset_augmentation}
\end{figure*}

\subsection{Predicting inter-class similarity across heterogeneous datasets}
In this section, we investigate the ability of \sysname on predicting the Inter-class similarities (Section~\ref{sec:scenarios}) for supplementing datasets with small number of samples with that of large number of samples.

To explore the potential of \sysname with regard to such questions, we conducted a set of pilot experiments. 
We consider $\mathbb{M}$ a new dataset, and $\{\mathbb{E}, \mathbb{R}, \mathbb{B}, \mathbb{F}\}$ the reference datasets. 
We train a 10-class MNIST classifier based on LeNet-5, but with as few as 10 samples per class from MNIST itself. 
We supplement each class of the training set with a varying number of samples borrowed from a third-party class that belongs to one of the reference datasets. 
Depending on the relevance of this third-party class to the MNIST class it is supplementing, it would be harmful or beneficial to the accuracy of our MNIST classifier. 
Our hypothesis is that \sysname-predicted similarity would help make informed pairings between third-party classes and MNIST classes in favorable ways. 
Formally, let $c_\mathbb{M}^i$ denote the $i$-th class of $\mathbb{M}$. We supplement $c_\mathbb{M}^i$ with the samples from the $k$-th class $c_Y^k$ of $Y$, where $Y \in \{\mathbb{E}, \mathbb{R}, \mathbb{B}, \mathbb{F}\}$.
We varied the number of original MNIST samples $n(c_\mathbb{M}^i)$ from 10 to 100. 
For each $n(c_\mathbb{M}^i)$, we varied the number of supplementing samples $n(c_Y^k)$ from 10 to 1000. 

To facilitate informed pairings between $c_\mathbb{M}^i$ and $c_Y^k$, we apply \sysname and two baseline methods for comparison, namely \textit{sample-MSE} and \textit{embeddings}. 
In \sysname, we pretrained per-class autoencoders $\mathcal{A}(c_Y^k)$ for all $Y\in\{\mathbb{E}, \mathbb{R}, \mathbb{B}, \mathbb{F}\}$ and all classes $c_Y^k$ thereof. 
Then the MNIST samples $c_\mathbb{M}^i$ are tested by all $\mathcal{A}(c_Y^k)$, forming a matrix of $\Delta(c_\mathbb{M}^i | c_Y^k)$ values for all $i$ and $k$ for each $Y$. 
The $(i,k)$ pair of the least $\Delta(c_\mathbb{M}^i | c_Y^k)$ decides the first pairing of $(c_\mathbb{M}^i, c_Y^k)$. Next pairs are iteratively decided in an increasing order of $\Delta(c_\mathbb{M}^i | c_Y^k)$ in the way that each $i$ and $k$ is chosen only once. 
In \textit{sample-MSE}, we estimate the similarity between all combinations of $(c_\mathbb{M}^i, c_Y^k)$ based on their mean Euclidean distances directly in the sample space, followed by the same pairing step. 
\textit{sample-MSE} is the baseline to compare \sysname-predicted similarity with those derived from the the sample space distances. 
In another baseline \textit{embeddings}, the samples of $c_\mathbb{M}^i$ and $c_Y^k$ are input to the per-dataset autoencoder $\mathcal{A}(Y)$.
Then we compute the euclidean distances between the `embeddings' of $c_\mathbb{M}^i$ and $c_Y^k$ that appear at the bottleneck of $\mathcal{A}(Y)$, followed by the same pairing step. 
We have \textit{embeddings} as a baseline in order to compare \sysname-predicted similarity with those derived from the embedding space distances. It is reported that embedding space distances often helps distill the semantic relevance~\cite{frome2013devise, mikolov2013efficient}, although it is controversial that a neural network model may be fooled to find short distances in embedding-space between visually unrelevent noisy inputs~\cite{nguyen2015deep}.

\begin{figure*}[ht]
    \centering
    \begin{subfigure}{0.245\textwidth}
        \centering
        \includegraphics[trim=1mm 0 12mm 6mm, clip=true, width=1.0\textwidth]{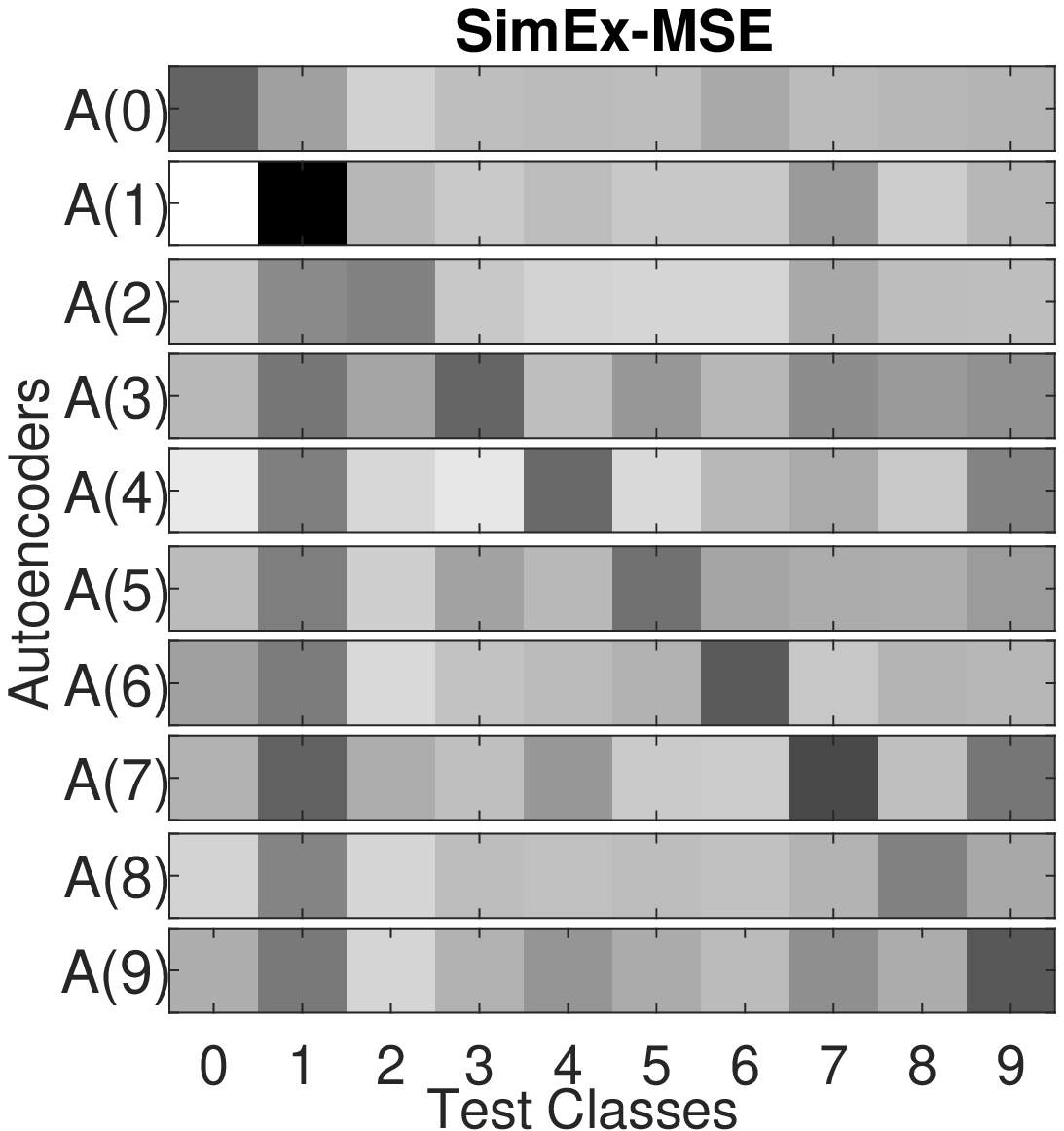}
        \label{fig:class_confusion_monomaths_mse}
    \end{subfigure}
    \begin{subfigure}{0.245\textwidth}
        \centering
        \includegraphics[trim=1mm 0 12mm 6mm, clip=true, width=1.0\textwidth]{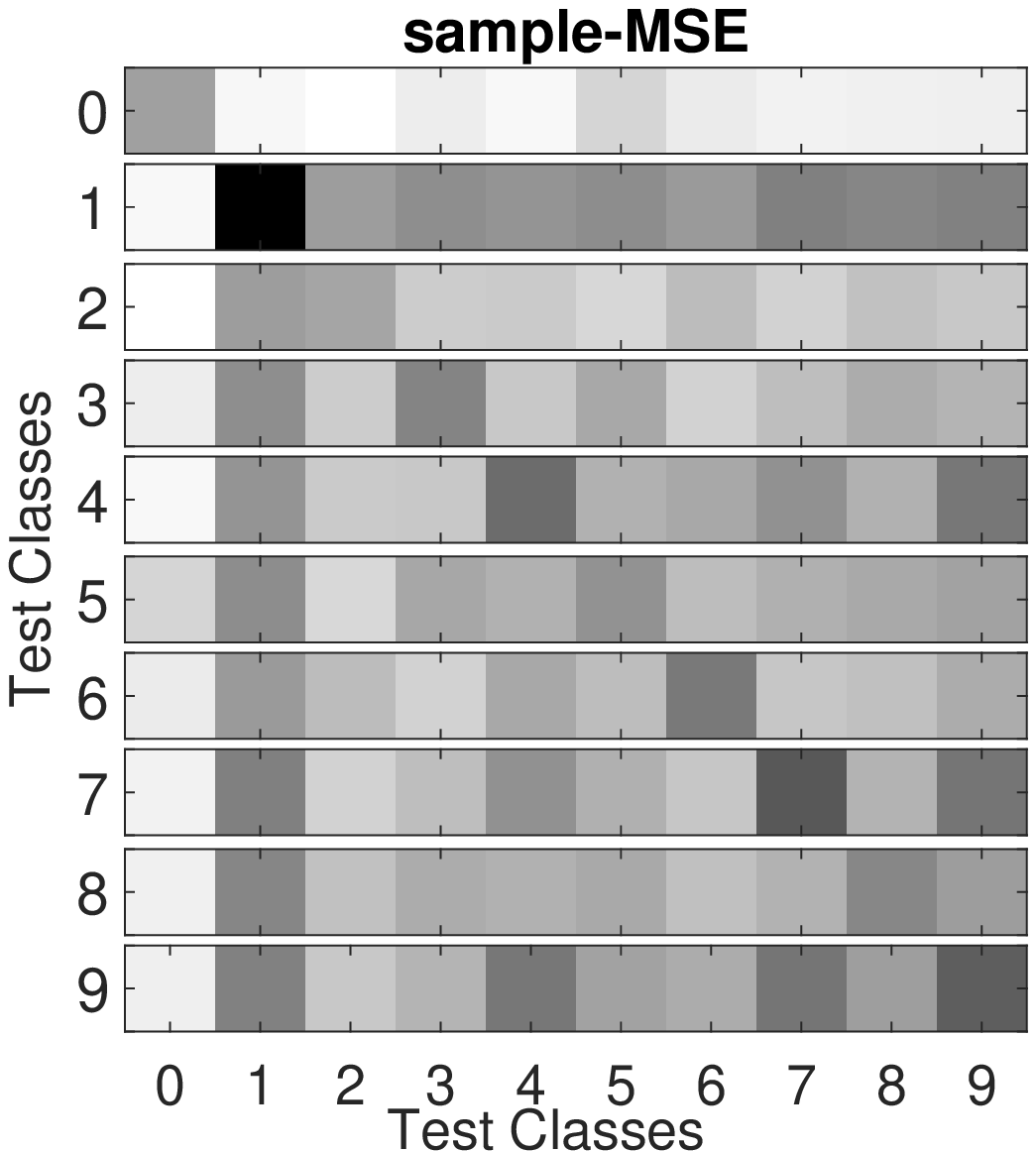}
        \label{fig:class_confusion_sample_mse}    
    \end{subfigure}
    \begin{subfigure}{0.245\textwidth}
        \centering
        \includegraphics[trim=1mm 0 12mm 6mm,, clip=true, width=1.0\textwidth]{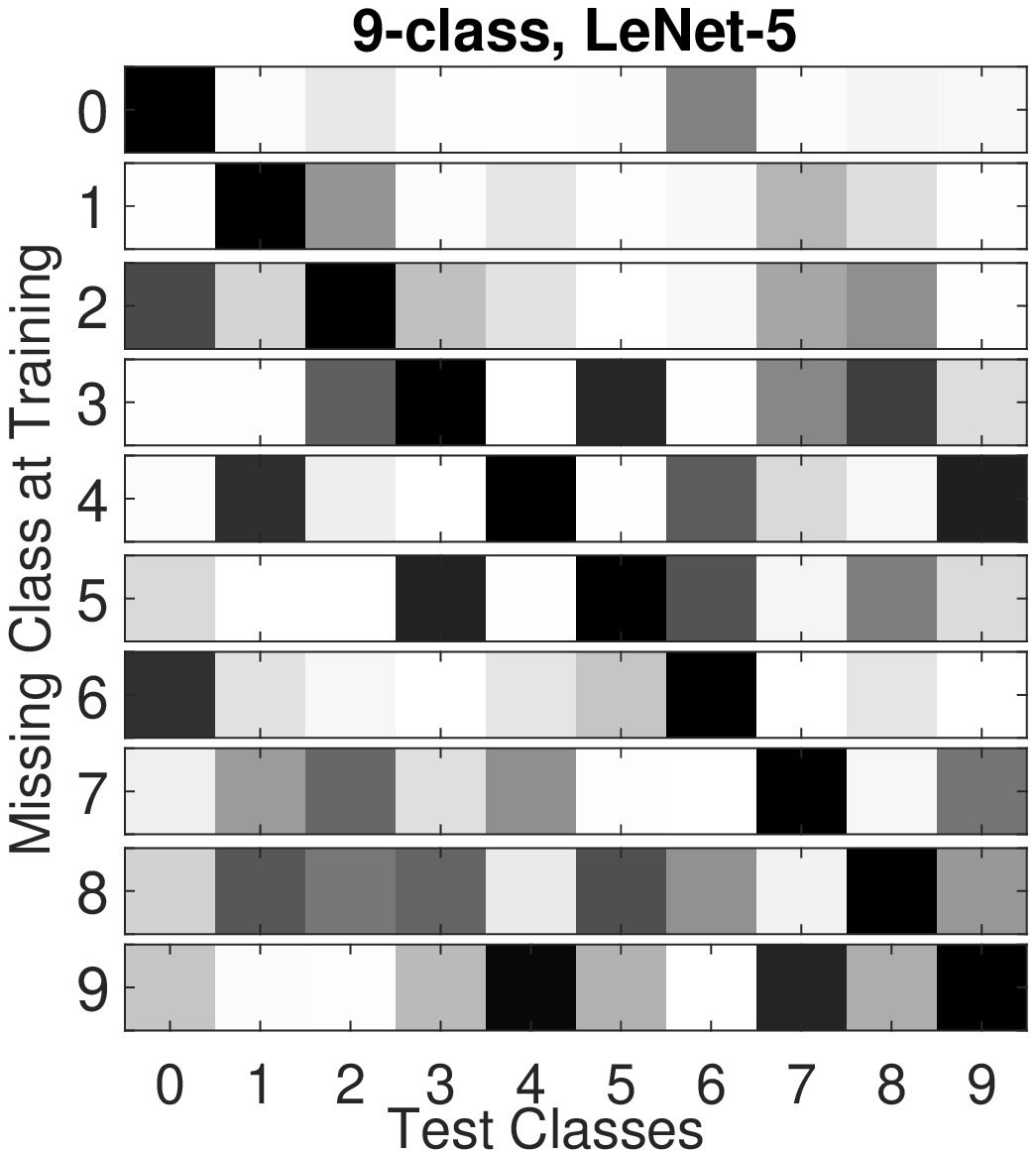}
        \label{fig:class_confusion_lenet5}
    \end{subfigure}
    \begin{subfigure}{0.245\textwidth}
        \centering
        \includegraphics[trim=1mm 0 12mm 6mm,, clip=true, width=1.0\textwidth]{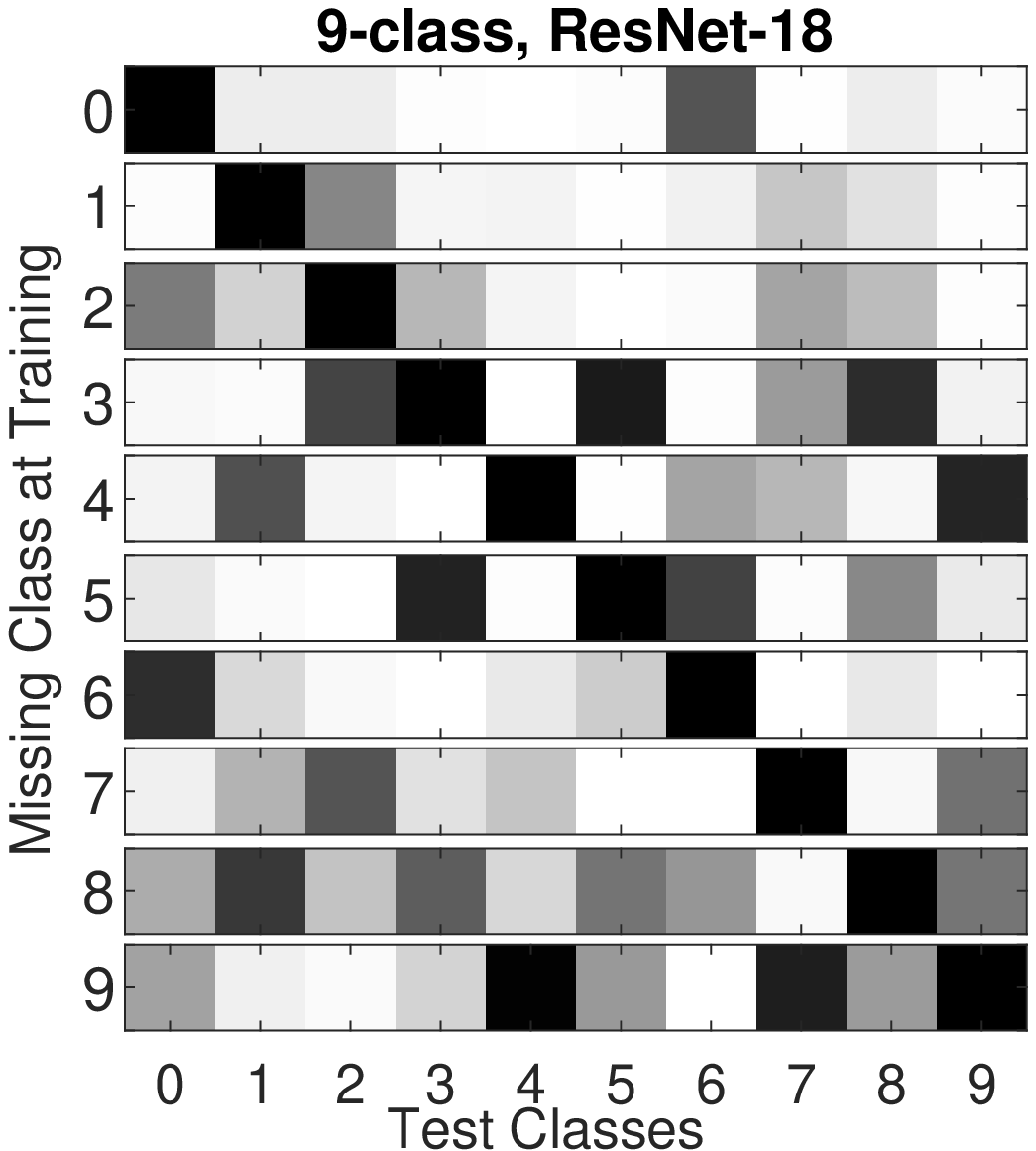}
        \label{fig:class_confusion_resnet18}    
    \end{subfigure}
    \caption{Inter-class similarity / confusion in MNIST dataset. Darker means more similar / confusing.}
    \label{fig:class_confusion}
\end{figure*}

Figure~\ref{fig:hetero_dataset_augmentation} enumerates the results of $4\times8\times4\times3 = 384$ combinations of \{$n(c_\mathbb{M}^i)$, $n(c_Y^k)$, $Y$, method\}. The first, second, and third rows represent \sysname, \textit{sample-MSE}, and \textit{embeddings} methods, respectively. 
The x-axes represent $n(c_Y^k)$, the number of supplementary training samples borrowed from the class $c_Y^k$ of the reference dataset $Y$, and the y-axes represent the 10-class MNIST classification accuracy at testing. 
The separate lines in each chart distinguish $n(c_\mathbb{M}^i)$, the number of original MNIST samples. 
The error bars represent the min and max accuracy out of 10 trials per combination. 
For testing, we used 1000 MNIST samples per class. 
Note that the \sysname results in the first row were obtained with MSE loss. Our \sysname experiments with iSSIM loss exhibited little difference from those with MSE loss here, thereby we omit the charts due to the page limit.

Several trends are visible. Not surprisingly, more MNIST samples available at training yield higher accuracy at testing. 
Borrowing more samples from third-party classes yield either (roughly) monotonically increasing or decreasing accuracy.
For $Y=\mathbb{B}$, it appears that \sysname made the more beneficial pairings compared to the two baselines under the same condition; adding more samples from $\mathbb{B}$ as informed by \sysname results in a larger boost of test accuracy. 
To reason, \sysname led to the pairings such that the $\mathbb{M}$ classes 0 through 9 correspond to the $\mathbb{B}$ classes 0 through 9 in the same order. However, \textit{sample-MSE} determined that $c_\mathbb{M}^1$ is closer to $c_\mathbb{B}^8$ than $c_\mathbb{B}^1$ and vice versa, resulting in the flipped pairings between the classes 1 and 8. 
\textit{embeddings} determined even more garbled pairings, resulting in further degraded accuracy.
For $Y=\mathbb{R}$, all three methods determined the identical pairings, resulting in similarly increasing accuracy. 
For $Y=\mathbb{E}$, the pairings determined by \sysname are neither convincingly beneficial nor harmful. However, the baselines exhibit slightly declining accuracy, implying that their pairings are less optimal. Detailed class-by-class pairings for EMNIST are listed in the supplementary materials. 

These experiments imply that \sysname could outperform the baseline methods in predicting the class-wise similarity between a new dataset and a given reference dataset, such that augmenting the new dataset's classes as per determined pairings is beneficial or less harmful to the original classification task, compared to the alternative pairings. 
A remaining question is how \sysname would provide informed knowledge to choose the right reference dataset out of many.
Note that, for two reference datasets $V$, $W$ and an unknown dataset $X$, $\Delta(X|V)$ and $\Delta(X|W)$ resulting from $\mathcal{A}(V)$ and $\mathcal{A}(W)$, respectively, may not be directly comparable to each other.
Especially when there is a large sample complexity difference between $V$ and $W$, it may result in a largely different dataset-wide error offsets at testing. 
$\mathbb{R}$ and $\mathbb{B}$ would be such a case.
As a preliminary attempt for regularization, we normalized the $\Delta(X|Y)$ values shown in Figure~\ref{fig:family_AE_mse_map} by the dataset-wide mean L2-norm of all samples for each $Y \in \{\mathbb{M}, \mathbb{E}, \mathbb{R}, \mathbb{B}, \mathbb{F}\}$. 
For example, $\Delta(\mathbb{M}|\mathbb{B})$, $\Delta(\mathbb{E}|\mathbb{B})$, ... , $\Delta(\mathbb{F}|\mathbb{B})$ are normalized by the mean L2-norm of all the samples $b_i$ ($\forall b_i \in \mathbb{B}$), and $\Delta(\mathbb{M}|\mathbb{R})$, $\Delta(\mathbb{E}|\mathbb{R})$, ... , $\Delta(\mathbb{F}|\mathbb{R})$ are by the mean L2-norm of $r_i$ ($\forall r_i \in \mathbb{R}$), and so on. Interestingly, the post-normalization order among all $\Delta(X|Y)$ where $X=\mathbb{M}$ was: $\Delta(\mathbb{M}|\mathbb{M}) < \Delta(\mathbb{M}|\mathbb{B}) < \Delta(\mathbb{M}|\mathbb{R}) < \Delta(\mathbb{M}|\mathbb{E}) < \Delta(\mathbb{M}|\mathbb{F})$, which happens to be  identical to the reverse order of positive accuracy boosts by \sysname exhibited in Figure~\ref{fig:hetero_dataset_augmentation}, i.e., $\mathbb{B}>\mathbb{R}>\mathbb{E}>\mathbb{F}$. 
Still this is by no means conclusive, but it is an interesting observation worth investigating further.

\subsection{Predicting inter-class similarity within a single dataset}


In this experiment, we explore the usability of \sysname-predicted similarity for early screening of inter-class confusion in a given dataset. 
We apply \sysname and sample-space distances for inter-class similarity (Section~\ref{sec:scenarios}) prediction. 
\sysname methods are analogous to Section~\ref{sec:methods} such that 10 autoencoders are trained, one per MNIST class, denoted $\mathcal{A}(0), \mathcal{A}(1), ..., \mathcal{A}(9)$. 
We train two sets of such autoencoders, one with MSE (denoted \textit{\sysname-MSE}) and the other with iSSIM (denoted \textit{\sysname-iSSIM}). 
The sample-space distances are the baselines.
For each pair of classes, mean sample distances are measured by two metrics, each denoted \textit{sample-MSE} and \textit{sample-iSSIM}, respectively.

We compare the inter-class similarity predicted by these methods against the confusion levels observed from MNIST classifiers. 
To observe the confusion levels, we attempted to adopt the distribution of MNIST classifier's outputs. 
However, training a MNIST classifier converges very quickly; even after a few epochs, testing samples $s_k$ from the \textit{k}-th class produces insignificantly small values at non-\textit{k} class output neurons. 
For a reliable alternative, we trained ten of 9-class classifiers, denoted $C_k$ for $0 \leq k \leq 9$. 
The training set of each $C_k$ is missing $s_k$, the samples of class $k$. 
At testing, $s_k$ yields nontrivial confusion levels from $C_k$, which the other methods' similarity predictions are compared against.
To mitigate possible model biases, we use two sets of 9-class classifiers from LeNet-5 and ResNet-18, respectively. These methods are denoted \textit{9class-Le5} and \textit{9class-Res18}, respectively.

\begin{table}[ht]
    \caption{Mean of Spearman's $\rho$ between inter-class similarity/confusion results by different methods}
    \centering
    \begin{small}
    \begin{tabular}{lp{1cm}p{1cm}p{1cm}p{1cm}}
        \toprule
                    & sample\newline -MSE & sample\newline -iSSIM & 9class\newline -Le5 & 9class\newline -Res18 \\ 
        \midrule
        \sysname-MSE    &    0.692            &         0.725         &     0.639           &     0.662            \\
        \sysname-iSSIM  &    0.765            &         0.804         &     0.629           &     0.646            \\               
        \midrule
        sample-MSE     &     n/a          &         n/a           &         0.438       &        0.442         \\
        sample-iSSIM   &     n/a          &         n/a           &         0.325       &        0.338         \\ 
        \bottomrule
    \end{tabular}
    \end{small}
    \label{tab:inter_class_spearman_rho}

\end{table}


Figure~\ref{fig:class_confusion} depicts the similarity / confusion results. We found that the results from \textit{\sysname-iSSIM} and \textit{sample-iSSIM} exhibit the trends very similar to those from \textit{\sysname-MSE} and \textit{sample-MSE}, thereby omitting their results. 
The different contrasts of the colormaps are not necessarily an issue as we are interested in their relative orders.
We assessed the consistency of similarity / confusion orderings in the same manner as in Section~\ref{sec:predicting_inter_dataset}. 
Table~\ref{tab:inter_class_spearman_rho}(left) lists the mean values of Spearman's $\rho$ between \textit{\sysname-}s and \textit{9class-}s, as well as between \textit{\sysname-}s and the baseline \textit{sample-}s. Given $\rho \in [-1, 1]$, the similarity predicted by both \textit{\sysname-} methods are of reasonably high consistency with the confusion observed from both \textit{9class-}s, as well as with the similarity predicted by the baseline \textit{sample-}s to equivalent extent. 
These results might seem that \textit{\sysname-} methods are no better than \textit{sample-}s. 
But Table~\ref{tab:inter_class_spearman_rho}(right) lists the $\rho$ values between \textit{sample-}s and \textit{9class-}s; \sysname is noticeably outperforming the baselines. 

These results indicate that the similarity orderings from \sysname lie at a middle point between those from \textit{sample-}s and \textit{9class-}s. The main implication is that \sysname may be more beneficial than the plain sample-space distances (even if the sample-space distance takes account of the structural similarity) in terms of predicting the potential confusion levels of an unknown class with respect to known classes of a dataset, although more experiments are necessary to warrant a strong claim.

\section{Concluding Remarks}

In this paper, we proposed \sysname, and experimentally explored the usability and benefits of \sysname-predicted similarity in the context of typical transfer learning and data augmentation. We demonstrated that \sysname-predicted data similarity is highly correlated with a number of performance differences mainly resulting from a dataset selection, e.g., in pretrained network transfer, train set augmentation, and inter-class confusion problems. We also showed that the \sysname's data similarity prediction exhibits equivalent or outperforming results compared to the baseline data comparisons in sample- or embedding-spaces. Importantly, we demonstrated \sysname achieving more than 10 times speed up in predicting inter-dataset similarity compared to the conventional transfer learning-based methods,as comparisons in \sysname require no training but inferences. 

Our pilot results have shown to the community the early potentials that this newly developed method could be usable and advantageous to several exemplary exercises in machine learning problems, especially those with a tight latency bound or a scalability requirement. We believe that further theoretical or empirical studies would guide us to a firm and deeper understanding on this newly developed method. 



\bibliography{references}
\bibliographystyle{sysml2019}

\end{document}